\documentclass{article}
\usepackage[table]{xcolor}
\usepackage[final]{corl_2026}
\usepackage{times}
\usepackage{amsmath,mathtools,amssymb}
\usepackage{graphicx}
\usepackage{booktabs,array,tabularx,multirow,makecell}
\usepackage{caption,subcaption,wrapfig}
\usepackage{pifont,xspace}
\usepackage[ruled,vlined]{algorithm2e}
\usepackage{cleveref}
\definecolor{iccvblue}{rgb}{0.21,0.49,0.74}
\usepackage{hyperref}

\newif\ifcrhighlight
\ifdefined\CRHighlight\crhighlighttrue\fi
\ifcrhighlight\hypersetup{hidelinks}\fi
\newcommand{\cradded}[1]{\begingroup\ifcrhighlight\color{blue}\fi#1\endgroup}
\title{DexNinja: Learning Robust Dexterous Cutting Policy with a Real-to-Sim-to-Real Data Engine}
\author{
Haozhe Lou$^{1,*}$, Runyi Yang$^{2,*}$, Wanqi Zhong$^{1}$, Caiyun Liu$^{3}$,\\
\bfseries Yicheng Liu$^{4}$, Wenhao Liu$^{5}$, Wenlong Ma$^{6}$, Jiawei Xia$^{7}$, Kanzhi Wu$^{8}$,\\
\bfseries Danda Pani Paudel$^{2}$, Luc Van Gool$^{2}$, Hang Zhao$^{4}$, Yiming Li$^{1}$\\[0.5ex]
\normalfont\footnotesize $^{1}$College of AI, Tsinghua University\quad $^{2}$INSAIT, Sofia University ``St. Kliment Ohridski''\\
\normalfont\footnotesize $^{3}$Peking University\quad $^{4}$IIIS, Tsinghua University\quad $^{5}$University of Southern California\\
\normalfont\footnotesize $^{6}$Beijing Institute of Technology\quad $^{7}$University of California, Berkeley\\
\normalfont\footnotesize $^{8}$vivo Mobile Communication Co., Ltd.\quad $^{*}$Equal contribution
}

\begin{document}
\raggedbottom
\maketitle

\begin{center}
    \vspace{-1cm}
    \includegraphics[width=0.72\linewidth]{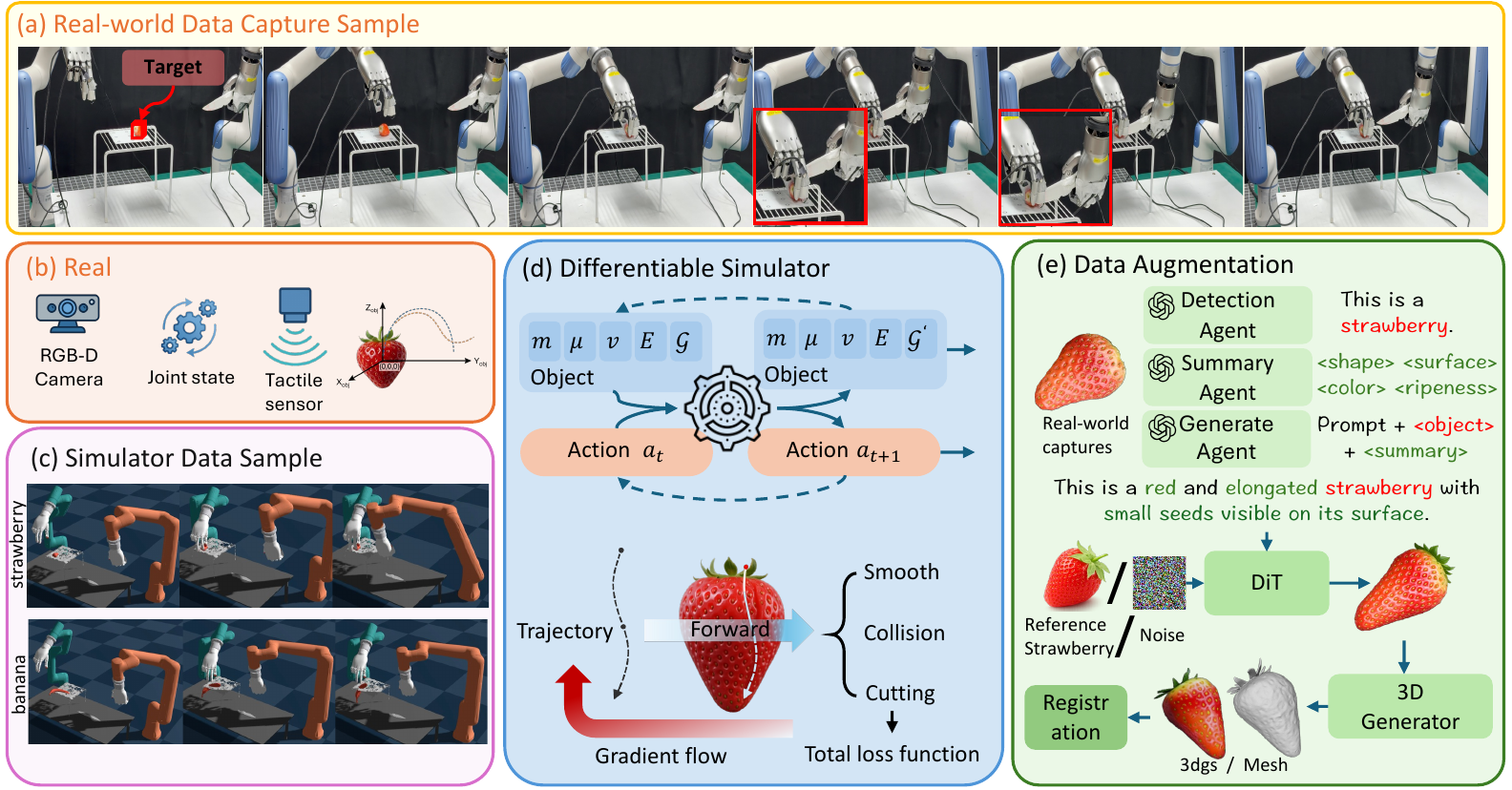}
    \captionsetup{hypcap=false}
    \captionof{figure}{\textbf{Real2Sim data densification pipeline.} (a) Real-world demonstration sample via teleoperation. (b) Real-world multi-modal sensing, and register object into a unified coordinate frame. (c) Simulation Data example (d) Differentiable simulator to generate trajectory. (e) Detection and summary agents extract structured object knowledge, which guides generative agents to produce diverse, in-domain instances to get controllable 3D generation.}
\label{fig:teaser}
\end{center}
\begin{abstract}
    Cutting a piece of food looks effortless in human hands. For a robot, however, every cut is a high-contact interaction with a deformable, elasto-plastic object whose behavior changes with pose, contact mode, and cutting stage. Collecting real-world data is expensive and wasteful: each trial typically contaminates or damages the food, so obtaining scalable real world data for dexterous cutting is expensive and difficult. Simulation is an appealing alternative, but a large sim-to-real gap exists in both physics and perception. This gap is especially severe for topology-changing interactions like cutting, where small errors in contact, friction, or material and elasticity responses quickly compound. We propose \textsc{DexNinja}, a real2sim2real framework that turns a handful of real demonstrations into a dense and realistic training distribution. \textsc{DexNinja} (i) reconstructs object instances from real captures, (ii) randomizes physically meaningful parameters under category-level constraints, and (iii) produces augmented dexterous manipulation episodes using a custom differentiable simulator that couples robot dynamics, tactile contact, and deformable cutting. We evaluate \textsc{DexNinja} on a dexterous food-slicing task and show that (1) with tens of real demonstrations and hundreds of augmented simulator data, the success rate could reach more than 60\%. (2) the augmented data consistently improves success rates when only using real data with more than 20\% via sim-to-real, and (3) the resulting policies generalize to out-of-distribution objects with varying shapes and sizes. The project page is at \href{https://dex-ninja.github.io/}{\textbf{https://dex-ninja.github.io/}}.
\end{abstract}

\section{Introduction}

Robotic manipulation has recently made strong progress for everyday household tasks such as table cleaning, dish washing, and picking up the trash~\cite{kim2024openvla,intelligence2025pi_,intelligence2025pi}. However, practical household autonomy requires also transform objects and materials in purposeful ways, like preparing food. Cooking involves deformable materials, dynamic contacts, and long-horizon tool-object interactions. \textit{Fruit cutting} is a representative example: the robot must stabilize an elasto-plastic ingredient while inducing controlled deformation and fracture with a knife. These tasks depend on contact history, friction, geometry, and material properties, making analytic heuristics difficult to generalize~\cite{wang2020effects,rico2007extending,han2020vision,mu2019robotic}. Prior cutting systems often rely on specialized tools or hardware~\cite{xu2023roboninja}, which can limit reuse across tools and tasks. Dexterous hands offer a promising alternative~\cite{shaw2023videodex,wang2022dexgraspnet}, providing richer stabilization and finer force modulation for contact-rich cooking tasks.


Actions like slicing a fruit demand rapid, continuous decisions about tool placement, object stabilization, and force control. Much of these competences comes from a tight perception-action loop that combines vision and touch: tactile cues reveal micro-slips, sticking events, and the propagation of a cut. \textit{Fruit cutting} is an unforgiving testbed for this closed-loop physical reasoning: deformable elasto-plastic objects, intermittent contacts, and topology changes as the object splits~\cite{jiang2016material}. These properties make \textit{fruit cutting} a representative task and a natural testbed for learning-based methods that must operate under irreversible state changes. Learning from real demonstrations is the most direct path to acquiring such behaviors, but for cutting it is unusually expensive and wasteful. Each trial damages the fruit, thus after a small number of executions the object becomes unsuitable for repeated collection. Consequently, real datasets are sparse and biased: they cover few object instances, and they concentrate around limited stages of the cutting process, leaving large portions of the state space underexplored. Simulation is an attractive alternative, but food deformation amplifies the sim-to-real gap. Success depends on accurate contact forces, frictional conditions, and deformable material behavior, while perception must remain stable as geometry and topology evolve~\cite{10.1145/3197517.3201293}. Small mismatches in physical parameters or sensing can derail the policy~\cite{jiang2016material,li2025pinwmlearningphysicsinformedworld,11127955}.

In this paper, we propose a \textbf{real-to-sim-to-real} (real2sim2real) strategy~\cite{lou2026d} to bridge this gap for topology-changing manipulation. Our key idea is to use \emph{category-level structure} as a scaffold for diversity: given a category label (e.g., ``strawberry''), we reconstruct object instances from sparse real demonstrations, then generate a family in the same category of physically plausible variations in a simulation process, while meaningfully diversifying geometry, mass, friction, and appearance. Thus we densify training distributions without drifting into unrealistic simulation artifacts. Crucially, differentiability couples reconstruction, parameter estimation, and simulation in a single loop, allowing gradients to guide policy learning. This design targets are difficult to obtain with either sparse real data or pure simulation alone: \textbf{Sim2Real through realistic densification:} starting from limited real demonstrations, we generate dense training data that remains close to real-world behavior, improving sim2real transfer across multiple learning baselines. \textbf{Real2Sim Generalization within a category:} by training on physically plausible category-level variations, we learn policies that remain effective on unseen objects with substantial shifts in physical parameters. Our contributions are:
\begin{enumerate}
    \vspace{-0.5em}
    \item \textbf{A real2sim2real data augmentation framework} that reconstructs object instances from sparse real demonstrations and perturbs geometry, physics, and appearance under category-level constraints to produce dense yet realistic training distributions.
    \vspace{-0.5em}
  \item \textbf{A learning framework for multi-modal, topology-changing manipulation} that leverages physically guided gradients for visual--tactile policy learning on deformable objects, moving beyond rigid-only assumptions in sim2real transfer.
  \vspace{-0.5em}
\end{enumerate}

\section{Related Works}
\vspace{-0.5em}
\subsection{Data generation and differentiable simulation}
\vspace{-0.5em}

Scalable data generation is increasingly studied for robot manipulation~\cite{fan2025twinalignervisualdynamicalignmentempowers,zhao2025real2edit2realgeneratingroboticdemonstrations,yang2025noveldemonstrationgenerationgaussian}. Trajectory-centric methods such as MimicGen generate large-scale simulation trajectories~\citep{mandlekar2023mimicgen}, while DemoGen extends this capability to real-world demonstrations~\citep{xue2025demogen}. R2RGen samples over object pose distributions~\citep{xu2025r2rgen}, and Giga-0 further expands variation to appearance and motion~\citep{gigaworldteam2025gigaworld0worldmodelsdata,yu2025real2render2real}. PointWorld directly synthesizes 3D point flows as motion representations~\cite{huang2026pointworld}. In contrast, we focus on \emph{physical densification} for topology-changing tasks: rather than only generating trajectories or poses, we densify \emph{mass}, \emph{geometry}, and \emph{friction} (optionally appearance) in a differentiable loop to improve system identification and policy robustness. Gaussian Splatting has become practical for robotics perception~\cite{kerbl20233d,li2025scenesplat, ma2025scenesplat++, li2025chorus,Li2025} and simulation~\cite{zheng2024gaussiangrasper}. Robo-GS~\cite{lou2024robogsphysicsconsistentspatialtemporal}, SplatSim~\cite{qureshi2025splatsim} and GSWorld~\cite{jiang2025gsworld}, unified Gaussian splatting for simulation-ready setting~\cite{jiang2025gsworld}. These works motivate appearance randomization and controllable rendering to reduce sim2real gaps~\cite{ye2024gaustudiomodularframework3d}. In our framework, appearance augmentation complements physics densification and can be integrated into the same pipeline~\cite{10260554,mittal2023orbit,turpin2022graspddifferentiablecontactrichgrasp,NEURIPS2018_842424a1,xu2022accelerated,Patel2022}. System identification recovers latent physical properties by matching predicted and observed dynamics, including inferring parameters from vision~\citep{pmlr-v78-standley17a} or videos~\citep{asenov2020vid2parammodellingdynamicsparameters}.

For topology-changing manipulation specifically, we build on physics-based simulators and differentiable cutting models such as Taccel~\citep{li2025taccel} and DiSECt~\citep{heiden2021disect}, which enable gradient-based simulation of fracture and topology change. RoboNinja studies cutting under randomized object properties~\citep{xu2023roboninja,lin2022diffskillskillabstractiondifferentiable}. Rule-based cutting plans tool paths and interaction forces along predefined trajectories~\citep{10246024}, while vision-based cutting incorporates visual feedback for alignment and deformation control~\citep{han2020vision} and force-aware controllers adapt motions using force sensing~\citep{wijayarathne2023real}. We treat mesh geometry as a initial condition for MPM based cutting dynamics, enabling structured densification around real objects~\citep{xian2023fluidlab,zhu2025one,zheng2024differentiable,chen2022daxbench}. Unlike pure randomization or heuristic control, we enforce topology consistency and anchor the distribution to real2sim reconstructed meshes or Gaussian splatting.

\subsection{Dexterous  manipulation and Teleoperation}

Dexterous manipulation spans universal grasping, task-specific skills, learning from video, and reward learning. Universal grasping enables zero-shot, contact-aware control~\cite{zhang2025robustdexgrasprobustdexterousgrasping,lin2024learningvisuotactileskillsmultifingered}, while task-specific methods learn skills such as in-hand rotation and compliant grasping~\cite{qi2022hand,chen2024springgraspsynthesizingcompliantdexterous,qi2023general,andrychowicz2020learning}. Learning from human videos reduces robot data needs but is often open-loop~\cite{shaw2023videodex}. Representation- and theory-driven works seek unified interaction abstractions for cross-task and cross-embodiment generalization~\cite{wei2025mathcaldrograspunifiedrepresentation,radosavovic2023real,nair2022r3m,shaw2023videodex,bahl2023affordances,xu2024flow}, while reward-based methods learn transferable visual rewards~\cite{ma2022vip}.

Tactile sensing is key to robust dexterous manipulation, especially for visual–tactile cutting where spatio-temporal force–contact cues matter~\cite{heng2025vitacformerlearningcrossmodalrepresentation}. We leverage differentiable tactile simulation to improve learning and generalization~\citep{huang2025spatiallyanchoredtactileawareness,10260554,si2024difftactilephysicsbaseddifferentiabletactile}. To obtain physically executable real2sim data, we retarget teleoperation demonstrations into the kinematic constraints. This enables efficient collection of contact-rich trajectories with  operator perception, while UMI further standardize and scale cross-embodiment datasets that support training generalist manipulation policies~\citep{sen2024learninglookaroundenhancing,chi2024universal,fastumi,liu2025rdt1bdiffusionfoundationmodel}.

\vspace{-1em}
\section{Problem Formulation}
\vspace{-1em}
\label{sec:problem formulation}
We consider the problem of \emph{fruit cutting}, in which one robotic arm equipped with a dexterous hand holds a knife, while the other arm uses its hand to interact with deformable food items.
At each discrete time step $t \in \{1,\ldots,T\}$, where $T$ denotes the horizon of the full episodes, the robot receives a multi-modal observation
\begin{equation}
\label{eq:observation}
  \mathbf{o}_t = \big\{ \mathbf{o}^{\text{vis}}_t,\ \mathbf{o}^{\text{tac}}_t,\ \mathbf{o}^{\text{prop}}_t \big\} \in \mathcal{O},
\end{equation}
where:
(i) $\mathbf{o}^{\text{vis}}_t$ denotes the third-view mounted RGB-D Camera, represented as a point cloud of the scene including the target instance fruit, tools and background,
(ii) $\mathbf{o}^{\text{tac}}_t$ contains tactile measurements from fingertip sensor arrays, and
(iii) $\mathbf{o}^{\text{prop}}_t$ denotes robot proprioceptive signals such as joint positions, joint velocities, and hand configuration states.

We represent each instance instantiated with physically interpretable parameters
\begin{equation}
\vartheta_t = \big(m,\mu_f,E,\nu,\mathcal{G}(t)\big),
\label{eq:instance_params}
\end{equation}
where $m$ is mass, $\mu_f$ is an effective Coulomb friction coefficient, $\mathcal{G}$ is the geometry (mesh/particles) regarding time, and $(E,\nu)$ are Young's modulus and Poisson's ratio for isotropic linear elasticity.
In simulation, we model 3D isotropic elasticity by converting Young's modulus $E$ and Poisson's ratio $\nu$ to the Lam\'e parameters:
\begin{equation}
\lambda = \frac{E\nu}{(1+\nu)(1-2\nu)}, \qquad
G = \frac{E}{2(1+\nu)},
\end{equation}
where $\lambda$ controls the volumetric response, and $G$, the shear modulus, controls shear resistance~\cite{xie2023physgaussian}. We treat the knife separately as a known rigid tool with a structured geometric model. The visual scene, excluding the knife model, is directly represented as point primitives:
\begin{equation} 
\mathbf{o}^{\text{vis}}_t = \{p_k\}_{k=1}^{N_P}, \qquad 
p_k \in\mathbb{R}^3
\end{equation}
where each point primitive contains a 3D position  obtained by reprojecting the RGB-D stream.

\begin{wrapfigure}{r}{0.48\linewidth}
    \vspace{-0.8em}
    \centering
    \includegraphics[width=\linewidth]{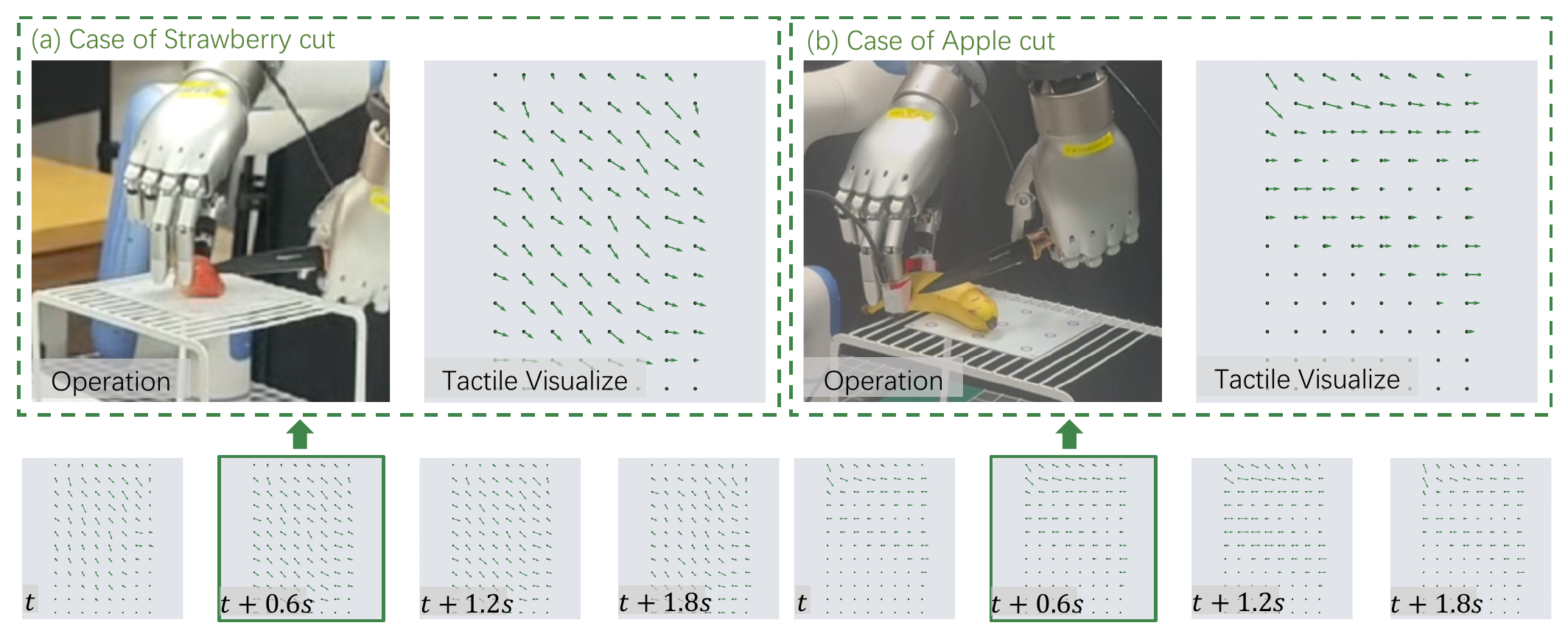}
    \caption{Tactile perception visualization of representative food-cutting examples.}
    \label{fig:tactile_perception}
    \vspace{-2.0em}
\end{wrapfigure}

The tactile observation $\mathbf{o}^{\text{tac}}_t$ encodes distributed contact measurements from fingertip tactile sensor arrays, $
\mathbf{o}^{\text{tac}}_t = \{ \mathbf{U}_t^{(f)} \}_{f=1}^{N_f}, 
\mathbf{U}_t^{(f)} \in \mathbb{R}^{H \times W \times C},
$ where $N_f$ is the number of tactile-equipped fingers, and each $\mathbf{U}_t^{(f)}$ is a spatial taxel array of resolution $H \times W$. Each taxel stores a $C$-dimensional local contact vector as the example seen in ~\cref{fig:tactile_perception}. For a taxel at grid location $(i,j)$, $\mathbf{u}_{ij} = \big( p_{ij},\ \boldsymbol{\gamma}_{ij},\ \delta_{ij} \big),$ where $p_{ij}$ denotes normal pressure, $\boldsymbol{\gamma}_{ij} \in \mathbb{R}^2$ represents tangential shear stress, and $\delta_{ij}$ is the local surface deformation.

The proprioceptive observation $\mathbf{o}^{\text{prop}}_t$ describes the internal state of the robot kinematics. We represent it as $
\mathbf{o}^{\text{prop}}_t = 
\big( \mathbf{q}_t,\ \dot{\mathbf{q}}_t,\ \mathbf{h}_t \big),
$ where $\mathbf{q}_t$ and $\dot{\mathbf{q}}_t$ denote the joint positions and joint velocities of the robot arm, respectively, and $\mathbf{h}_t$ represents hand configuration states, including finger joint angles.

Let $\mathbf{o}_t \in \mathcal{O}$ be the observation at time $t$ from~\eqref{eq:observation}. 
We decompose the robot action as
\begin{equation}
a_t = \big(a_t^{\text{kin}},\, a_t^{\text{tac}}\big),
\qquad
a_t^{\text{kin}} \in \mathcal{A}^{\text{kin}},\;\;
a_t^{\text{tac}} \in \mathcal{A}^{\text{tac}},
\end{equation}
where $a_t^{\text{kin}}$ controls large-scale kinematic motions (left and right arm), while $a_t^{\text{tac}}$ models contact-sensitive hand control. Conditioned on the static visual geometry, we model the kinematic component as a mapping $f_\phi(\cdot)$ conditioned on visual observation,
which implemented in practice via object-centric motion specification that maintains the desired blade--fruit geometry. The tactile component is modeled as a tactile and proprioceptive observation conditioned feedback policy $\pi_\theta(\cdot)$ which is closed-loop to tactile and proprioceptive observation.
We note a dataset of each episode of expert demonstrations
$
\mathcal{D} = 
\Big\{
\big(\mathbf{o}_t,\, a_t\big)
\Big\}.$
Each tuple consists of the deployable observation $\mathbf{o}_t \in \mathcal{O}$, the executed action $a_t \in \mathcal{A}$. 
We learn a hybrid control architecture with two parameterized policies: a kinematic policy $f_{\phi}$ and a tactile policy $\pi_{\theta}$. Given demonstration data \(\mathcal{D}\), we optimize
\begin{equation}
\begin{aligned}
\hat a_t^{\text{kin}}&
=
f_\phi\!\left(\mathbf{o}_t^{\text{vis}}\right), \quad
\hat a_t^{\text{tac}}=
\pi_\theta\!\left(
\mathbf{o}_{0:t}^{\text{tac}},
\mathbf{o}_{0:t}^{\text{prop}}
\right),\\
\mathcal{L}\!\left(\hat a_t,a_t\right)
&=
\mathcal{L}_{\text{kin}}\!\left(
\hat a_t^{\text{kin}},a_t^{\text{kin}}
\right)
+
\mathcal{L}_{\text{tac}}\!\left(
\hat a_t^{\text{tac}},a_t^{\text{tac}}
\right).
\end{aligned}
\label{eq:bc_objective}
\end{equation}

\section{Real2Sim2Real pipeline: Dexninja}
\label{sec:pipeline}

We propose DexNinja, a real2sim2real pipeline for dexterous elasto-plastic \textit{fruit cutting}~\cite{xu2023roboninja,Genesis,lou2024robogsphysicsconsistentspatialtemporal,pfaff2025scalablereal2simphysicsawareasset,sundaresan2022diffcloud}. Stage I: From each teleoperated real demonstration, DexNinja reconstructs a digital food asset. Stage II: DexNinja uses a differentiable cutting simulator to generate physically consistent training episodes~\cite{ze2025twist,ze2025twist2,joao2025gmr,qin2023anyteleop}. Then in the simulator, the fruit diversity is densified by the digital assets generation pipeline. Stage III : The policy $f_\phi$ and $\pi_\theta$ is trained with aligned vision, tactile, and proprioceptive observations. You can find the implementation detail in the supplementary.

\textbf{Stage I:} We first teleoperate a successful cutting trial and record synchronized visual, tactile, and proprioceptive trajectories. From this demonstration, we reconstruct the food geometry using mesh reconstruction~\cite{lou2024robogsphysicsconsistentspatialtemporal} and estimate the food pose with FoundationPose~\cite{foundationposewen2024}. We then convert the recorded observation into a physically grounded training sample by aligning all reconstructed assets at metric scale, registering them to an object-centric template in canonical space, and estimating a fixed end-effector--knife transform from the teleoperation logs.

\textbf{Stage II:} We initialize the reconstructed assets in a deformable cutting simulator based on Roboninja and Warp~\cite{warp,xu2023roboninja,Genesis,geng2025roboverse}. For each rollout, we sample the food geometry and physical parameters, instantiate an object-centric knife-motion template, and apply local perturbations such as starting-position offsets, approach-angle changes, and small pose noise. The resulting knife path is converted into executable end-effector targets. We then use IK~\cite{sundaralingam2023curobo} to solve feasible joint-space trajectories from these targets, execute the trajectories in simulation, and record temporally aligned visual, tactile, proprioceptive, and action data. The simulator models deformable and topology-changing food dynamics, capturing contact evolution, material separation, and tactile feedback during slicing.

To further densify the training set, we generate additional in-category food assets through the digital generation pipeline described in Sec.~\ref{sec:data augmentation}. This yields an augmented dataset
\(\mathcal{D}_{\text{aug}}\)
that combines the real demonstration with physically consistent simulated rollouts. The differentiable trajectory objective used for rollout filtering and refinement is provided in Supplementary.

When we register the canonical object--knife trajectory into the real-world hardware setting, for bi-manual execution, we synchronize both arms through a shared timeline so that the stabilizing grasp precedes cutting and maintains contact throughout the interaction. To ensure safety and feasibility, we perform collision-checking and trajectory filtering using fast motion planning libraries ~\cite{Genesis,sundaralingam2023curobo}. We reject trajectories that violate self-collision constraints, collide with the environment (table, branches, etc.), or approach kinematic singularities. Finally, we bridge residual sim2real discrepancies with targeted randomization at  fruit pose noise.

\begin{figure*}[htbp]
    \centering
    \includegraphics[width=1.\linewidth]{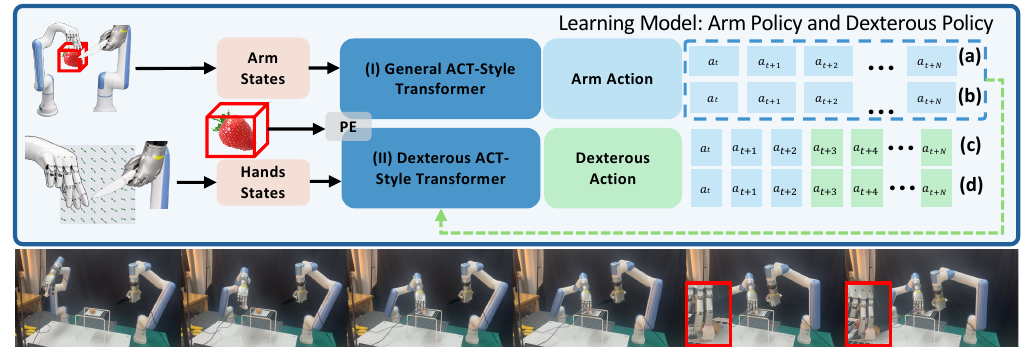}
    \caption{ Learning policy to control arm and dexterous hand separately via adaptive action chunking. We give an real-world inference result at bottom.}
    \label{fig:pipeline}
\end{figure*}

\textbf{Stage III:} On the curated and augmented dataset \(\mathcal{D}_{\text{aug}}\), we train an ACT-style policy~\cite{zhao2023learningfinegrainedbimanualmanipulation} that maps multimodal observations to real-world action chunks. We factor the action prediction into arm and hand components: the arm controls global approach and pose alignment, while the hand handles local grasp and contact regulation. Tactile inputs are encoded as effector-specific token groups associated with the corresponding hand or knife, improving robustness to localized tactile transients~\cite{wen2026grdextertechnicalreport}. The policy predicts a stage-conditioned action chunk:
\begin{equation}
\hat a_{t:t+H_t-1}^{\text{kin}}
=
f_\phi\!\left(\mathbf{o}_t^{\text{vis}}, z_t\right), \quad
\hat a_{t:t+H_t-1}^{\text{tac}}=
\pi_\theta\!\left(
\mathbf{o}_{t}^{\text{tac}},
\mathbf{o}_{t}^{\text{prop}},
z_t
\right),
\end{equation}
where \(z_t\in\mathcal{Z}\) denotes the current manipulation stage.

To improve Sim2Real robustness, we use variable action chunking during post-training~\cite{so2025improvinggenerativebehaviorcloning,weng2025temporalactionselectionaction}. Demonstration anchors define the stage token \(z_t\) and a nominal horizon \(\bar{H}(z_t)\). We sample the execution horizon around this nominal value:
\begin{equation}
H_t \sim p(H \mid z_t)
= \lambda\, \mathrm{Unif}\!\big(H_{\min}(z_t), \bar{H}(z_t)\big) 
+ (1-\lambda)\, \mathrm{Unif}\!\big(\bar{H}(z_t), H_{\max}(z_t)\big).
\end{equation}
Shorter chunks are used for contact-sensitive stages, while longer chunks are used for coarse free-space motion. During post-training, the policy is supervised on expert chunks of length \(H_t\). At inference, the selected chunk is executed with correponding chunk size
$a_{t:t+H_t-1} \leftarrow \pi_\theta(\mathbf{o}_t, z_t)$. In our real system, we use three stages with nominal horizons:
\textbf{approach \& grasp} \((\bar{H}=64)\),
\textbf{knife positioning} \((\bar{H}=128)\), and
\textbf{cutting with tactile feedback} \((\bar{H}=16)\).

\section{Continuation on a Shape Manifold for Data Augmentation}
\label{sec:data augmentation}

\label{subsec:densification}

In Sec~\ref{sec:pipeline}, we detailed the full pipeline. And in this section, we will explain the data augmentation pipeline specifically. We model each food category as a continuous family of shapes induced by a learned geometric prior. Starting from a nominal real instance for which a cutting motion succeeds, we densify training data by (i) sampling nearby instances on the category shape family and (ii) transferring the nominal motion to these nearby instances via local continuation.

\begin{wrapfigure}{r}{0.65\linewidth}
    \centering
    \includegraphics[width=0.95\linewidth]{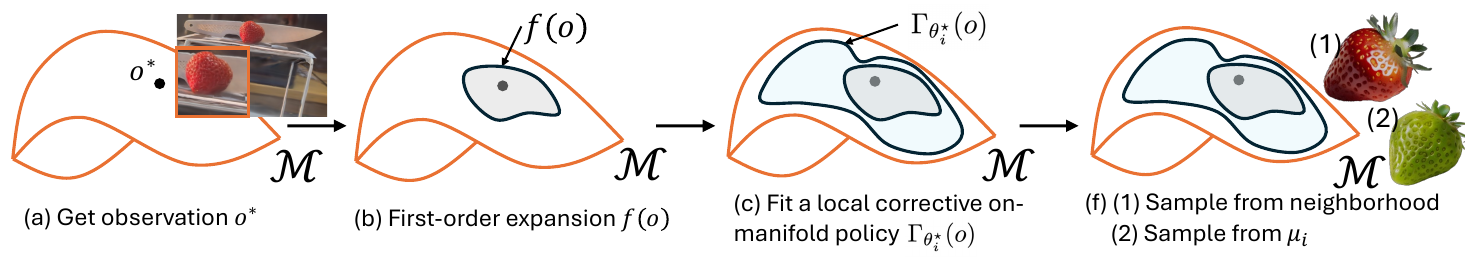}
    \vspace{-0.5em}
    \caption{On-manifold approximation of real-data neighborhood for data augmentation pipeline.}
    \label{fig:data_densification}
    \vspace{-1em}
\end{wrapfigure}
\paragraph{Data augmentation pipeline}
Fig.~\ref{fig:teaser}(e) summarizes the full generation setting.
Given real-captured references, we first infer the object category using a detector.
A summarization module then extracts attribute-level descriptors (e.g., coarse shape, surface characteristics) and encodes them into a structured semantic representation designed to remain consistent across domains.
Conditioned on the predicted category and these attributes, a generative module synthesizes diverse yet physically plausible instances that remain within the target domain distribution.
This knowledge-guided generation enables controlled variation in geometry and appearance while preserving task-relevant physical properties, yielding large-scale, domain-consistent data densification.
The resulting in-domain textual descriptions are used to generate controllable images, which are then lifted to 3D assets.
Additional implementation details are provided in the supplementary material.
This pipeline induces a learned shape prior and yields a nominal reconstructed instance, which we formalize next.

\paragraph{Shape generation and local manifold structure}
Let $\mathcal{Z}\subset\mathbb{R}^{d_z}$ denote the latent space and let
\begin{equation}
\mathcal{G}:\mathcal{Z} \rightarrow \mathcal{M},
\qquad
z \mapsto \mathcal{G}(z),
\label{eq:shape_generator}
\end{equation}
be an implicit geometry generator, where $\mathcal{M}$ is the family of physically plausible food geometries represented by the implicit model.
We use the term \emph{shape manifold} in the local sense that, under an appropriate geometry metric $d_{\mathcal{M}}(\cdot,\cdot)$, the generator is locally regular: small perturbations in $z$ induce small changes in geometry.\footnote{Concretely, it suffices that $\mathcal{G}$ is locally Lipschitz: for $z$ in a neighborhood of $z_0$, $d_{\mathcal{M}}(\mathcal{G}(z),\mathcal{G}(z_0)) \le L_{\mathcal{G}}\|z-z_0\|$.}
Given a reconstructed real instance, we obtain a nominal latent code $z_0$ and geometry $\mathcal{G}(z_0)$.
We densify the training distribution by sampling $z$ within a neighborhood of $z_0$, thereby producing a continuum of related instances (Fig.~\ref{fig:data_densification}).

\paragraph{Local regularity and analytical reliability}
Our continuation argument relies on local smoothness of the mapping from deployable visual observations to actions,
$f(\mathbf{o}^{vis})$.
Assuming $f$ is differentiable in a neighborhood of a successful anchor observation $o^\star$, a first-order expansion gives
\begin{equation}
f(o) \;=\; f(o^\star) \;+\; J_f(o^\star)(o-o^\star) \;+\; r(o),
\label{eq:taylor_f}
\end{equation}
where $J_f(o^\star)$ is the Jacobian and the remainder satisfies $
\lim_{\mathbf{o}\to \mathbf{o}^\star}
\|r(\mathbf{o})\| /
\|\mathbf{o}-\mathbf{o}^\star\|
=0.$
Equivalently, by the mean-value theorem, there exists a neighborhood $\mathcal{N}(o^\star)$ and a constant $L_f$ such that
$\|f(o)-f(o^\star)\|\le L_f\|o-o^\star\|$ for all $o\in\mathcal{N}(o^\star)$.
This formalizes the key intuition: small instance perturbations admit proportionally small action corrections, which is precisely the regime in which continuation is numerically stable and analytically justified.

\paragraph{Neighborhood of demonstrations}
Let $\mathcal{D}_0=\{(o_i,a_i)\}_{i=1}^{N_0}$ denote deployable observation--action pairs collected from a teleoperated episode.
We evaluate candidate actions in simulation using a binary feasibility predicate $S(o,a)\in{0,1},$ which indicates whether the action succeeds under both local (object-level) and global (scene-level) constraints, i.e., it passes the cutting and collision tests.

We capture this neighborhood by fitting a local corrective policy $\Gamma_{\theta_i}$ around each anchor:
\begin{equation}
\theta_i^\star
\;=\;
\arg\max_{\theta}\;
\mathbb{E}_{o \sim \mu_i}
\Big[
S\!\big(o,\Gamma_{\theta}(o)\big)
\;-\;
\beta \,\|\Gamma_{\theta}(o)-a_i\|^2
\Big],
\label{eq:ancoda_obj_rewrite}
\end{equation}
where $\mu_i$ is a distribution over observations induced by
(i) small state perturbations,
(ii) physics randomization, and
(iii) local shape variation $z\sim \nu_i$ through \eqref{eq:shape_generator}.
The quadratic term anchors the correction to the demonstrated action $a_i$, while allowing the policy to compensate for mismatches introduced by deformation and partial separation during cutting.
In this sense, $\Gamma_{\theta_i^\star}$ acts as a \emph{continuation operator}: it tracks a successful behavior as instance parameters vary locally along the learned shape family, while remaining close to the nominal action to ensure numerical stability.

\paragraph{Augmented data generation}
We generate additional samples by rolling out the local corrective policies in simulation and retaining only successful outcomes:
\begin{equation}
\mathcal{D}_{\text{aug}}
=
\mathcal{D}_0
\;\cup\;
\bigcup_{i=1}^{N_0}
\Big\{(o,a)\;:\; o\sim \mu_i,\; a=\Gamma_{\theta_i^\star}(o),\; S(o,a)=1\Big\}.
\label{eq:ancoda_aug_rewrite}
\end{equation}
Compared to naive perturbation, continuation concentrates samples on regions that remain physically feasible under geometry variation and cutting-induced topology changes.

A single teleoperated episode captures only a narrow slice of the skill (one instance, one grasp, one approach).
We therefore use simulation as a data densification engine (Fig.~\ref{fig:data_densification}) to expand this slice into a broad set of consistent rollouts over diverse instances sampled from the category prior, while restricting observations to deployable sensing modalities.
\cradded{\paragraph{Real-world anchoring and admission}
Unlike independent domain randomization, our category-conditioned prior anchors variation to real captures. A category is defined by shape, surface, colour, and ripeness attributes spanning the collected commercial varieties (Supplementary Sec.~4). Here, ``plausible'' is an operational admission rule: assets must pass semantic/multi-view consistency, metric-scale, geometry-validity, simulator-compatibility, and stable-initialization checks. Rollouts randomize category-bounded $\theta=(m,\mu,E,\nu)$, refine trajectories for the sampled instances, and must pass IK, collision, and cutting-success checks.\par}

\vspace{-1em}
\section{Experimental Evaluation}
\vspace{-1em}
\label{sec:experiment}

We experiment on a real food cutting task and deploy the policy on-robot under real contact conditions. We structure the evaluation around two questions: \textbf{Sim-to-real transfer:} Can the policy execute meaningful cutting behaviors in real-world setups? \textbf{Generalization:} Does simulation-based data densification improve performance, on unseen instances within the same food category?

\subsection{Task setup and evaluation protocol}
\label{subsec:task_setup}






\newcommand{\best}[1]{\textbf{#1}}
\newcommand{\pp}[1]{\underline{#1}} 

\begin{wraptable}{R}{0.45\textwidth}

  \centering
  \small
  \setlength{\tabcolsep}{4pt}
  \renewcommand{\arraystretch}{1.12}

  \resizebox{\linewidth}{!}{%
  \begin{tabular}{@{}l l c c c@{}}
    \toprule
    \makecell[l]{\textbf{Training}\\\textbf{regime}} &
    \textbf{Food} &
    \textbf{Progress}~$\uparrow$ &
    \textbf{Success}~$\uparrow$ &
    \textbf{Integrity}~$\uparrow$ \\
    \midrule

    \multirow{5}{*}{\textbf{Sim-only}}
      & Strawberry & 10/25 & 6/25 & \pp{10/25} \\
      & Apple      & 5/25  & \pp{4/25} & \pp{5/25} \\
      & Banana     & 16/25 & \pp{12/25} & \pp{14/25} \\
      & Potato     & \pp{15/25} & \best{12/25} & \pp{11/25} \\
      & \textbf{Avg.} & \pp{46/100} & \pp{34/100} & \pp{40/100} \\
    \midrule

    \multirow{5}{*}{\textbf{Real-only}}
      & Strawberry & \pp{12/25} & \pp{8/25} & \pp{10/25} \\
      & Apple      & \pp{6/25}  & \pp{4/25} & 4/25 \\
      & Banana     & \pp{17/25} & \pp{12/25} & \pp{14/25} \\
      & Potato     & 9/25 & \pp{7/25} & 9/25 \\
      & \textbf{Avg.} & 44/100 & 31/100 & 37/100 \\
    \midrule

    \multirow{5}{*}{\makecell[l]{\textbf{Sim+Real}\\\textbf{(Ours)}}}
      & Strawberry & \best{19/25} & \best{12/25} & \best{14/25} \\
      & Apple      & \best{14/25} & \best{8/25}  & \best{7/25}  \\
      & Banana     & \best{24/25} & \best{18/25} & \best{23/25} \\
      & Potato     & \best{19/25} & \best{12/25} & \best{16/25} \\
      & \textbf{Avg.} & \best{76/100} & \best{50/100} & \best{60/100} \\
    \bottomrule
  \end{tabular}%
  }
   
  \caption{Sim-to-real gap and the effect of mixing simulation and real demonstrations, with 25 trials per food.}
  \vspace{-2em}
  \label{tab:sim2real_densification_results}
  
\end{wraptable}

The main results are trained and tested on 4 food types: \textit{strawberry, apple, banana, potato}. In the training, for each instance, we adopt 5 demonstrations for real-world and 50 sequences from simulation. In supplementary, we show the results of training-set scaling-up. For each type, we select $25$ instances per category for evaluation. Natural variation in geometry  and physical properties is treated as uncontrolled but bounded variability representative of real deployment. We consider a food-halving task following~\cite{wijayarathne2023real}, where a robot must cut a food item into two halves without causing crumbling. A trial is successful only if the knife fully traverses the object and the halves can be cleanly separated. To assess robustness to shifts in initial conditions, we randomize object placement across trials. Specifically, the food item is placed on the table with a random planar offset relative to the robot base frame within a predefined reachable workspace region. We do not pre-assume or pre-calibrate strawberry shape. We report 3 metrics, each measured as the number of trials satisfying the criterion out of $25$: \textbf{Progress ($\uparrow$):} whether the policy can approach the food with the knife and reach a valid pre-cut state, i.e., the knife successfully reaches the target region and establishes an initial cutting contact with the surface. \textbf{Success ($\uparrow$):} whether the knife cuts through the food, i.e., the cut fully traverses the item, yielding two separable halves and ending with a safe termination/retreat. \textbf{Integrity ($\uparrow$):} whether the food remains in acceptable condition after the attempt.
\cradded{Integrity is judged manually as a binary heuristic requiring no obvious unintended crushing, mashing, tearing, excessive squeezing, or fragment loss.}


\subsection{Sim-to-real gap and the effect of mixing sim and real data}
\label{subsec:sim2real_gap}

To quantify the sim2real gap and evaluate the benefit of densification, we train policies under 3 data regimes: \textbf{Sim-only:} simulated cutting trajectories only. \textbf{Real-only:} teleoperated real demonstrations only. \textbf{Sim+Real (Augmented):} real demonstrations combined with augmented simulation data. All policies are evaluated on the same held-out real-world setups, which controls for evaluation conditions and isolates the effect of training data. This protocol also directly tests whether densification improves within-category generalization to unseen instances. In total, we conduct 300 real-world experiments across different data regimes and different instances.

As shown in Tab.~\ref{tab:sim2real_densification_results}, policies trained on either sim-only or real-only data generalize poorly to unseen real-world setups. Aggregated over 100 real trials (4 foods × 25 instances), Sim-only achieves 46\% progress, 34\% success, and 40\% integrity, while Real-only achieves 44\% progress, 31\% success, and 37\% integrity. In contrast, Sim+Real (Augmented) achieves the strongest performance, reaching 76\% progress, 50\% success, and 60\% integrity.
Overall, co-training with densified simulation improves progress, success, and integrity compared to the two single-source baselines, indicating that our augmentation pipeline mitigates the sim-to-real gap and improves within-category generalization. Additional experimental results are provided in the supplementary material.

\subsection{Complex Motion and Heterogeneous Objects}

\begin{figure}
    \centering
    \includegraphics[width=0.95\linewidth]{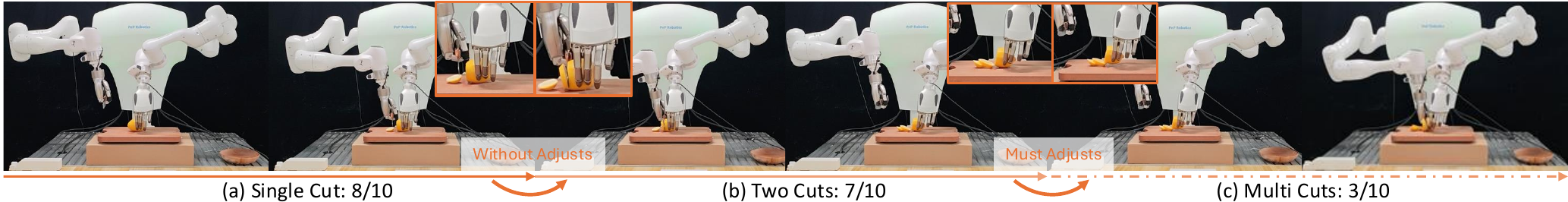}
    \caption{Real-world lemon cutting results under different cutting requirements. The success rate is declining as the manipulation complexity increase.}
    \label{fig:multi_cuts_quali}
\end{figure}
We further evaluate our method on \textit{lemons}, which present more complex deformation and heterogeneous internal structure. This setting is used to test the ability of the policy to handle multi-cut behavior on non-uniform objects, the training details are consistent with Tab.~\ref{tab:sim2real_densification_results}. Our method is not based on a simple forceful cutting strategy; instead, it explicitly models object rigidity and leverages simulation to capture the interaction between the knife and the object. As shown in Fig.~\ref{fig:multi_cuts_quali}, the policy achieves a high success rate for single-cut trials, with 8 successes out of 10 attempts. Performance decreases as the number of required cuts increases, achieving 7/10 for two cuts and 3/10 for multi-cut trials. This trend indicates that while the proposed method can handle heterogeneous objects and repeated cutting motions to some extent, while complex multi-stage cutting remains more challenging due to accumulated errors, increased object deformation, more difficult contact dynamics and more dexterous adjust steps. Additional ablation results are provided in the supplementary material.

\vspace{-1em}

\cradded{\subsection{Component ablations and failure analysis}
\label{subsec:rebuttal_analysis}
Each variant follows the 5-real/50-simulated-sequence protocol of Table~\ref{tab:sim2real_densification_results}, with 25 held-out fruits per category across four categories (100 trials). Table~\ref{tab:rebuttal_ablations} isolates quality control, physical randomization, trajectory refinement, tactile observation, and generated assets. Removing trajectory refinement causes the largest drops in Progress and Success; removing tactile observation most strongly affects Integrity. Reconstruction-only training reduces geometric coverage. These matched-budget ablations support the contribution of the components beyond sample count alone.\par}
\begin{table}[t]
\centering
{%
\captionsetup{font=small}
\caption{Component ablations. $\Delta$ is relative to Full DexNinja.}
\label{tab:rebuttal_ablations}
\footnotesize
\setlength{\tabcolsep}{4pt}
\begin{tabularx}{\linewidth}{@{}Xcrrr@{}}
\toprule
Variant & P/S/I & $\Delta$P & $\Delta$S & $\Delta$I \\
\midrule
Full DexNinja (Table~\ref{tab:sim2real_densification_results}) & 76/50/60 & -- & -- & -- \\
w/o Quality Control & 72/46/55 & $-4$ & $-4$ & $-5$ \\
w/o Physical Randomization & 72/44/51 & $-4$ & $-6$ & $-9$ \\
w/o Trajectory Refinement & 65/39/53 & $-11$ & $-11$ & $-7$ \\
w/o Tactile Observation & 73/43/45 & $-3$ & $-7$ & $-15$ \\
w/o Generated Assets (Reconstruction Only) & 67/42/50 & $-9$ & $-8$ & $-10$ \\
\bottomrule
\end{tabularx}}
\end{table}
\begin{table}[t]
\begin{minipage}[t]{0.48\linewidth}
\centering
{%
\caption{Failed-trial analysis.}
\label{tab:rebuttal_failures}
\footnotesize
\setlength{\tabcolsep}{3pt}
\begin{tabularx}{\linewidth}{@{}Xr@{}}
\toprule
Cause & Trials \\
\midrule
Perception/registration & 8 \\
Grasp/contact control & 19 \\
Cutting-physics mismatch & 17 \\
Kinematic/safety/collision & 6 \\
\bottomrule
\end{tabularx}}
\end{minipage}\hfill
\begin{minipage}[t]{0.49\linewidth}
\centering
{%
\caption{Quality check of assets.}
\label{tab:rebuttal_qc}
\footnotesize
\setlength{\tabcolsep}{2pt}
\begin{tabularx}{\linewidth}{@{}Xrrrr@{}}
\toprule
Category & Proposed & 2D QC & 3D QC & Accepted \\
\midrule
Strawberry & 397 & 361 & 326 & 300 \\
Apple & 376 & 349 & 321 & 300 \\
Banana & 453 & 407 & 354 & 300 \\
Potato & 444 & 396 & 343 & 300 \\
Clean total & 1,670 & 1,513 & 1,344 & 1,200 \\
\bottomrule
\end{tabularx}}
\end{minipage}
\end{table}

\cradded{Table~\ref{tab:rebuttal_failures} assigns each of the 50 failed Sim+Real trials to its earliest cause using synchronized videos and logs. Grasp/contact control and cutting-physics mismatch are the main failure sources. A cutting-physics mismatch denotes stable registration, grasp, and trajectory tracking but a material response not captured by simulation, such as blade stalling or fruit compression instead of penetration. Table~\ref{tab:rebuttal_qc} reports the asset-generation funnel from 1,670 proposals to 1,200 accepted assets.\par}


\section{Conclusion}
\label{sec:conclusion}

We presented DexNinja, a real2sim2real data augmentation framework for dexterous food cutting.
From a small set of teleoperated demonstrations, DexNinja reconstructs and registers object assets with multi-modal sensing, estimates physically interpretable parameters, and expands the training distribution via physics-aware randomization and knowledge-guided instance generation. A custom differentiable simulator couples robot dynamics, tactile contact, and deformable cutting for scalable rollouts and gradient-based fitting, while policies use an ACT-style transformer with separated arm/hand control and adaptive
action chunking for robust execution. Experiments on various instances indicate that sim-only or real-only training is insufficient:
combining real demonstrations with DexNinja's augmented pipeline improves robustness to unseen shapes and sizes. 

\section{Limitations}
This work has two primary limitations. First, it relies on a mounted camera setup, which may limit deployment flexibility. We will deploy with wrist camera for future work. Second, the base policies considered in our experiments are object-specific. The focus of this paper is therefore on validating the effectiveness of data densification for complex dexterous manipulation.
\cradded{Current policies are category-specific and new categories require new assets and retraining or fine-tuning.}


\label{cr:main-end}
\clearpage
\acknowledgments{This research was partially funded by a collaboration project between VIVO and INSAIT on scene understanding and editing. Research was also partially funded by the Ministry of Education and Science of Bulgaria (support for INSAIT, part of the Bulgarian National Roadmap for Research Infrastructure).}
 
\bibliography{example}
\clearpage
\begingroup
\color{black}
\providecommand{\pp}[1]{\textbf{#1}}
\providecommand{\best}[1]{\textbf{#1}}
\setcounter{section}{0}
\setcounter{subsection}{0}
\setcounter{equation}{0}
\setcounter{table}{0}
\setcounter{figure}{0}
\setcounter{footnote}{0}
\renewcommand{\theHsection}{supp.\arabic{section}}
\renewcommand{\theHsubsection}{supp.\arabic{section}.\arabic{subsection}}
\renewcommand{\theHequation}{supp.\arabic{equation}}
\renewcommand{\theHtable}{supp.\arabic{table}}
\renewcommand{\theHfigure}{supp.\arabic{figure}}
\renewcommand{\theHfootnote}{supp.\arabic{footnote}}
\section*{Supplementary Material}
\label{cr:supplement-start}
\section{Simulation Details}
\label{sec:supp_sim_details}

This section describes how DexNinja converts real teleoperation demonstrations into physically consistent simulated rollouts. We denote coordinate frames by $\mathcal{F}_{(\cdot)}$ and rigid transforms by
${}^{A}\mathbf{T}_{B}\in SE(3)$, which maps a point expressed in frame $\mathcal{F}_B$ into frame $\mathcal{F}_A$.
We use frames: world $\mathcal{F}_W$, food $\mathcal{F}_S$, knife $\mathcal{F}_K$, and end-effector $\mathcal{F}_E$.
A time-parameterized knife trajectory is $\{{}^{W}\mathbf{T}_{K}(t)\}_{t=0}^{T}$, and the ending time is $t_f$.

\subsection{Stage I: Teleoperation and Object Frame}
\label{subsec:supp_retargeting}
We teleoperate the robot to perform one successful cutting trial and record multi-modal trajectories.
Each time step includes the observation $\mathbf{o}_t=\{\mathbf{o}^{\text{vis}}_t,\mathbf{o}^{\text{tac}}_t,\mathbf{o}^{\text{proprio}}_t\}$ and the executed action (joint angles).
We additionally estimate the food pose ${}^{W}\mathbf{T}_S$ and its geometry~\cite{lou2024robogsphysicsconsistentspatialtemporal,foundationposewen2024}. Then extracting the cutting trajectory in the food/object frame:
\begin{equation}
  {}^{S}\mathbf{T}_{K}(t) = \left({}^{W}\mathbf{T}_{S}\right)^{-1}\, {}^{W}\mathbf{T}_{K}(t),
  \qquad
  {}^{S}\mathbf{T}_{K,f} = {}^{S}\mathbf{T}_{K}(t_f).
  \label{eq:knife_in_strawberry_frame}
\end{equation}
The sequence \(\{{}^{S}\mathbf{T}_{K}(t)\}\) captures the demonstrated
cutting skill as a relative knife-object motion, which is transferable
across reconstructed food instances.

\subsection{Stage II: Differentiable Cutting Simulator Details}
\label{subsec:supp_traj_objective}

A single real episode provides a narrow slice of the state-action distribution. The goal of Stage~II is to turn this single slice into a broad and physically grounded training set.
We reconstruct the task assets (object, knife, support surface) in metric scale and refine relative poses so that the object-centric template in \cref{eq:knife_in_strawberry_frame} is consistent with the reconstructed geometry. When the knife is rigidly grasped, we estimate the constant transform from end-effector to knife via teleoperation logs:
\begin{equation}
  {}^{E}\mathbf{T}_{K}(t) = \left({}^{W}\mathbf{T}_{E}(t)\right)^{-1}\, {}^{W}\mathbf{T}_{K}(t),
\end{equation}
This estimate provides the fixed grasp geometry needed to convert object-centric knife motions into executable end-effector trajectories. We then initialize the reconstructed assets in our simulator ~\cite{xu2023roboninja,Genesis,geng2025roboverse} and enable deformable, topology-changing dynamics for the food. This is essential: cutting changes contact geometry and can induce separation, so a rigid-body proxy is insufficient for producing reliable feedback during slicing.
We sample strawberry instances by varying shape and physical parameters $\theta$ (e.g., mass, compliance, friction) and generate perturbed object-centric knife trajectories $\{{}^{S}\mathbf{T}_{K}^{(i)}(t)\}$ around the demonstrated template (start offsets, approach angles, small pose noise). For each sampled trajectory, we map the knife motion back to an end-effector motion using the estimated grasp geometry:
\begin{equation}
  {}^{W}\mathbf{T}_{E}^{(i)}(t)
  = {}^{W}\mathbf{T}_{S}\; {}^{S}\mathbf{T}_{K}^{(i)}(t)\; \left({}^{E}\mathbf{T}_{K}\right)^{-1}.
  \label{eq:ee_pose_from_object_centric_knife}
\end{equation}
We then solve inverse kinematic~\cite{sundaralingam2023curobo} to obtain feasible joint-space trajectories $\mathbf{q}^{(i)}(t)$ and execute them in simulation to generate augmented dataset $\mathcal{D} = \big\{\big(\mathbf{o}_t^{(i)}, a_t^{(i)}\big)\}$.

\paragraph{Trajectory generation loss.}
We optimize a knife trajectory $\{(\mathbf{x}_i,\theta_i)\}_{i=0}^{T}$ by differentiating through the unrolled simulator. Knife--ground interaction is evaluated with the ground signed distance field (SDF) $d(\mathbf{p})$. At each step we sample $K$ points across the blade width~\cite{xu2023roboninja,lin2022diffskillskillabstractiondifferentiable,mu2019robotic,10246024}
\begin{equation}
\mathbf{p}_{i,k}=\mathbf{x}_i+\alpha_k\,w_{\text{knife}}\,\mathbf{dir}(\theta_i),\quad
\alpha_k=\frac{k}{K-1}
\end{equation}
\begin{equation}
\mathbf{dir}(\theta_i)=\begin{bmatrix}-\sin\theta_i & \cos\theta_i & 0\end{bmatrix}^{\!\top}
\end{equation}

and use a smooth hinge $\mathrm{srelu}(x)=\tfrac12\!\left(x+\sqrt{x^2+\varepsilon}\right)$.

We define the nonnegative knife--ground gap
\begin{equation}
g_i=\frac{1}{K}\sum_{k=0}^{K-1}\mathrm{srelu}\!\big (d(\mathbf{p}_{i,k})\big).
\end{equation}
The cutting objective encourages approaching the surface and making stepwise progress:
\begin{equation}
\mathcal{L}_{\text{cut}}
=
\frac{1}{T+1}\sum_{i=0}^{T} w_i\, g_i^2
\;+\;
\frac{\lambda_{\text{prog}}}{T}\sum_{i=1}^{T}
\Big[\mathrm{srelu}\!\big(g_i-(1-\rho)g_{i-1}\big)\Big]^2,
\quad
\end{equation}
\begin{equation}
w_i=\left(\frac{i+1}{T+1}\right)^2.
\end{equation}

We use the convention $d(\mathbf{p})>0$ outside the ground, $d(\mathbf{p})=0$ on its surface, and $d(\mathbf{p})<0$ inside. Here $m\geq0$ is a small numerical penetration tolerance, not a positive clearance margin. The collision penalty smoothly approximates the fourth power of penetration beyond this tolerance, i.e., $d(\mathbf{p})<-m$:
\begin{equation}
\mathcal{L}_{\text{col}}
=
\frac{1}{(T+1)K}\sum_{i=0}^{T}\sum_{k=0}^{K-1}
\Big[\mathrm{srelu}\!\big(-(d(\mathbf{p}_{i,k})+m)\big)\Big]^4.
\end{equation}
We add smoothness/consistency and effort regularizers:
\begin{equation}
\mathcal{L}_{\text{rot}}=\frac{1}{T}\sum_{i=0}^{T-1}(\theta_{i+1}-\theta_i)^2,\quad
\mathcal{L}_{\text{move}}=\frac{1}{T+1}\sum_{i=0}^{T}(\theta_i^{k}-\theta_i^{v})^2
\end{equation}
and minimize the weighted sum
\begin{equation}
\mathcal{L}
=
w_{\text{cut}}\mathcal{L}_{\text{cut}}
+w_{\text{col}}\mathcal{L}_{\text{col}}
+w_{\text{rot}}\mathcal{L}_{\text{rot}}
+w_{\text{move}}\mathcal{L}_{\text{move}}
\end{equation}

\subsection{Tactile Simulation During Cutting}
\label{subsec:supp_tactile_cutting}
The right-hand data generation consists of two components. First, in the pre-grasp stage, we reuse expert grasp priors~\cite{zhang2024dexgraspnet20learninggenerative} to generate a kinematically feasible hand pose that encloses the strawberry. Second, during cutting, we simulate tactile feedback and use it to generate adaptation data for maintaining stable contact under object deformation.

We implement tactile sensing in our differentiable MPM-based simulator~\cite{xu2023roboninja}. For each fingertip or tactile pad, we construct a signed distance field (SDF) in the local sensor frame. At each simulation step, we identify food particles that enter the near-surface band of the SDF. Using the particle positions and velocities, we estimate contact events and compute contact impulses or forces from two quantities: the penetration depth and the relative tangential velocity between the particles and the sensor surface. We then aggregate the per-particle contact contributions into a compact tactile observation by spatially binning the forces over the sensor surface.

To generate tactile adaptation data, we simulate the full grasp-and-cut sequence starting from the initial grasp. During cutting, the deformable strawberry changes shape and contact distribution under the fingers. When the tactile signals indicate incipient slip, we apply small corrective finger motions to maintain stable contact while still allowing compliant deformation. We record the resulting state--action pairs as supervision for the second-stage right-hand policy.

For sim-to-real transfer, we synthesize tactile observations in a frame-consistent manner. Instead of expressing tactile signals in the world frame, we anchor contact locations and force directions to the active end-effector or tool frame. This representation uses visual pose estimates and calibrated extrinsics to align tactile observations with the current tool pose~\cite{huang2025spatiallyanchoredtactileawareness}. By reducing spurious variation caused by object placement and camera viewpoint, the frame-consistent tactile representation improves transfer across scenes and embodiments.

Finally, we calibrate a lightweight mapping from simulated forces to real tactile sensor readings, including scale, offset, and noise models. This calibration improves robustness to contact variations between simulation and real-world execution.

\subsection{Scene level and hardware registration}
\label{sec:sim2real_supp}

We register the canonical object--knife trajectory to the real-world hardware setting before execution. Given the estimated fruit pose in the world frame, we compute the rigid transform from the canonical object frame to the world frame and apply it to the nominal knife trajectory. The resulting world-frame knife targets are converted into end-effector targets using the estimated end-effector--knife transform, and then into robot joint trajectories through inverse kinematics and time-parameterization. During this conversion, we enforce joint limits, velocity and acceleration limits, and end-effector orientation constraints.

For bi-manual execution, both arms are synchronized through a shared timeline. The stabilizing hand first reaches the pre-grasp configuration and establishes contact with the fruit. The cutting arm then executes the registered knife trajectory while the stabilizing hand maintains contact throughout the interaction. To ensure feasibility and safety, we perform collision checking and trajectory filtering using fast motion-planning libraries~\cite{Genesis,sundaralingam2023curobo}. We reject trajectories that violate self-collision constraints, collide with the surrounding environment such as the table or branches, or approach kinematic singularities. Finally, to reduce residual sim-to-real discrepancies, we apply targeted randomization to the fruit pose during data generation. This complements the object--knife canonical planning and improves robustness when transferring the learned policy across different robot embodiments.

A practical challenge is that deformable fruit cutting is not natively supported in the Genesis backend used for scene-level robot simulation. In contrast, prior cutting simulators such as RoboNinja and DiSect model deformable cutting interactions, but do not provide the full articulated robot stack required for scene-level kinematics and dynamics~\cite{xu2023roboninja,heiden2021disect}. To combine these complementary capabilities, we introduce a \emph{local-to-scene registration} bridge that uses the knife as the shared interface between the two simulation layers.

Specifically, we run a local, tool-centric cutting simulation to resolve the tool--fruit interaction, including contact and tactile signals. The resulting interaction is then projected onto the knife as external contact information. We aggregate the local contact into contact points expressed in the knife frame, together with the corresponding force and moment applied to the knife. These quantities can also be summarized as an equivalent net wrench at the knife center of mass. We then register the resulting knife pose or trajectory to the scene-level simulator, where the articulated robot executes the same knife motion under robot kinematic constraints while receiving the mapped external wrench as disturbance or feedback.

This bridge avoids re-implementing deformable cutting inside a full scene-level robot engine. Although it introduces additional per-step data transfer between the local cutting simulator and the scene-level simulator, the design is modular and can be reused across different objects, tools, and robot embodiments. In practice, the minimal information passed across the bridge consists of the knife pose or trajectory $\{T_{\mathrm{knife}}(t)\}$, contact points $\{p_i(t)\}$, the knife center of mass $c(t)$ or a fixed tool COM, and the external forces and moments $\{f_i(t), \tau_i(t)\}$, or equivalently a net wrench.

\section{Training Details}
\label{sec:training_details}

DexNinja trains the deployment policy on the augmented dataset
$\mathcal{D}_{\mathrm{aug}}$ produced by the real2sim2real pipeline. Each sample follows the notation in the main paper:
\[
\mathbf{o}_t
=
\left\{
\mathbf{o}^{\mathrm{vis}}_t,
\mathbf{o}^{\mathrm{tac}}_t,
\mathbf{o}^{\mathrm{prop}}_t
\right\},
\qquad
a_t=
\left(
a^{\mathrm{kin}}_t,
a^{\mathrm{tac}}_t
\right).
\]
Here $a^{\mathrm{kin}}_t$ controls large-scale arm, wrist, and tool motion, while
$a^{\mathrm{tac}}_t$ controls contact-sensitive dexterous-hand behavior. The training section below gives the implementation details behind the stage-conditioned ACT-style policy used in the main paper.

\paragraph{Why a single globally normalized action head is unstable.}
Although both the arm and the dexterous hand are ultimately commanded in joint space, their action statistics are very different. Finger joints often traverse a large fraction of their physical range during a grasp or contact adjustment, whereas arm joints usually make smaller corrections for approach, alignment, and compliant knife motion. A single globally normalized action head therefore creates an ill-conditioned learning problem: either the high-amplitude hand actions dominate the loss and gradients, or the action scale is reduced to stabilize the arm and the hand becomes under-actuated. This mismatch becomes more severe after contact begins, because small pose bias, friction variation, and compliance errors can quickly lead to slip, over-compression, or failed cutting.

We therefore keep the main-paper decomposition $a_t=(a^{\mathrm{kin}}_t,a^{\mathrm{tac}}_t)$ and implement it with separate action heads and group-wise action normalization. The curriculum also follows the task structure: Stage~1 brings the system to a reliable pre-contact configuration, while Stage~2 uses tactile feedback to regulate contact and complete the cut.

\subsection{Stage-conditioned action chunking}
\label{subsec:supp_stage_conditioned_chunking}

Let
\[
\mathcal{D}_{\mathrm{aug}}
=
\left\{
\left(
\mathbf{o}^{(n)}_{1:T_n},
a^{(n)}_{1:T_n},
z^{(n)}_{1:T_n}
\right)
\right\}_{n=1}^{N}
\]
denote the training dataset, where $z_t\in\mathcal{Z}$ is the manipulation-stage token. In our implementation,
\[
\mathcal{Z}
=
\left\{
z^{\mathrm{app}},
z^{\mathrm{knife}},
z^{\mathrm{cut}}
\right\},
\]
corresponding to approach/grasp, knife positioning, and tactile-guided cutting.

Following the main paper, the policy predicts a stage-conditioned action chunk:
\begin{equation}
\hat{\tilde a}^{\mathrm{kin}}_{t:t+H_t-1}
=
f_{\phi}
\left(
\mathbf{o}^{\mathrm{vis}}_t,
z_t
\right),
\qquad
\hat{\tilde a}^{\mathrm{tac}}_{t:t+H_t-1}
=
\pi_{\theta}
\left(
\mathbf{o}^{\mathrm{tac}}_t,
\mathbf{o}^{\mathrm{prop}}_t,
z_t
\right),
\label{eq:supp_stage_conditioned_policy}
\end{equation}
where $\hat{\tilde a}$ denotes the normalized action output. In practice, $\mathbf{o}^{\mathrm{tac}}_t$ and $\mathbf{o}^{\mathrm{prop}}_t$ may contain a short temporal stack, but they retain the same deployable tactile and proprioceptive modalities as the main formulation.

The chunk horizon is sampled around a stage-dependent nominal horizon:
\begin{equation}
H_t
\sim
p(H\mid z_t)
=
\lambda_H
\,\mathrm{Unif}
\left(
H_{\min}(z_t),
\bar H(z_t)
\right)
+
(1-\lambda_H)
\,\mathrm{Unif}
\left(
\bar H(z_t),
H_{\max}(z_t)
\right).
\label{eq:supp_adaptive_chunk_sampling}
\end{equation}
We use longer chunks for free-space motion and shorter chunks for contact-rich control:
\[
\bar H(z^{\mathrm{app}})=64,
\qquad
\bar H(z^{\mathrm{knife}})=128,
\qquad
\bar H(z^{\mathrm{cut}})=16.
\]
This matches the design in the main paper: coarse approach and knife-positioning stages benefit from temporal smoothing, while tactile-guided cutting requires frequent replanning.

\subsection{Group-wise action normalization and execution}
\label{subsec:supp_groupwise_action_norm}

For each action group
$g\in\{\mathrm{kin},\mathrm{tac}\}$,
we compute group- and stage-specific statistics from the training set:
\[
\boldsymbol{\mu}_{g}(z),
\qquad
\boldsymbol{\sigma}_{g}(z).
\]
The supervised target used for training is
\begin{equation}
\tilde a^g_{\tau}
=
\operatorname{clip}
\left(
\frac{
a^g_{\tau}
-
\boldsymbol{\mu}_{g}(z_t)
}{
\boldsymbol{\sigma}_{g}(z_t)+\epsilon_a
},
-1,
1
\right),
\qquad
g\in\{\mathrm{kin},\mathrm{tac}\}.
\label{eq:supp_groupwise_norm}
\end{equation}
This avoids forcing arm-scale and hand-scale actions into a single global normalization range.

At execution time, normalized policy outputs are mapped to joint increments using group-specific step sizes:
\begin{equation}
\Delta \hat{\mathbf q}^{g}_{\tau}
=
\mathbf{s}^{g}(z_t)
\cdot
\operatorname{clip}
\left(
\hat{\tilde a}^{g}_{\tau},
-1,
1
\right),
\qquad
g\in\{\mathrm{kin},\mathrm{tac}\},
\label{eq:supp_groupwise_delta_q}
\end{equation}
where $\mathbf{s}^{\mathrm{kin}}$ is chosen to be smaller than $\mathbf{s}^{\mathrm{tac}}$ for fine arm/tool corrections near contact. The commanded joint state is then projected into the safe robot command set:
\begin{equation}
\mathbf q_{\tau+1}
=
\Pi_{\mathcal Q}
\left(
\mathbf q_{\tau}
+
\left[
\Delta \hat{\mathbf q}^{\mathrm{kin}}_{\tau},
\Delta \hat{\mathbf q}^{\mathrm{tac}}_{\tau}
\right]
\right),
\label{eq:supp_joint_update}
\end{equation}
where $\Pi_{\mathcal Q}$ enforces joint limits, rate limits, workspace bounds, and collision-related safety constraints.

\subsection{Chunked behavior-cloning objective}
\label{subsec:supp_chunked_bc_objective}

The main-paper behavior-cloning objective decomposes into kinematic and tactile losses. For action-chunk training, we apply the same decomposition over the sampled horizon $H_t$:
\begin{equation}
\mathcal{L}_{\mathrm{BC}}
=
\frac{1}{H_t}
\sum_{\ell=0}^{H_t-1}
\left[
\omega_{\mathrm{kin}}(z_t)
\left\|
\hat{\tilde a}^{\mathrm{kin}}_{t+\ell}
-
\tilde a^{\mathrm{kin}}_{t+\ell}
\right\|_1
+
\omega_{\mathrm{tac}}(z_t)
\left\|
\hat{\tilde a}^{\mathrm{tac}}_{t+\ell}
-
\tilde a^{\mathrm{tac}}_{t+\ell}
\right\|_1
\right].
\label{eq:supp_chunked_bc}
\end{equation}
The weights $\omega_{\mathrm{kin}}(z_t)$ and $\omega_{\mathrm{tac}}(z_t)$ emphasize the relevant action group for each stage. During approach and knife positioning, the kinematic loss receives larger weight; during tactile-guided cutting, the tactile loss receives larger weight while allowing low-rate kinematic correction.

We also regularize the executed joint increments:
\begin{equation}
\mathcal{L}_{\mathrm{rate}}
=
\frac{1}{H_t-1}
\sum_{\ell=0}^{H_t-2}
\left\|
\left[
\Delta \hat{\mathbf q}^{\mathrm{kin}}_{t+\ell+1},
\Delta \hat{\mathbf q}^{\mathrm{tac}}_{t+\ell+1}
\right]
-
\left[
\Delta \hat{\mathbf q}^{\mathrm{kin}}_{t+\ell},
\Delta \hat{\mathbf q}^{\mathrm{tac}}_{t+\ell}
\right]
\right\|_2^2.
\label{eq:supp_rate_loss}
\end{equation}

\subsection{Sim2Real deployment}
\label{subsec:supp_sim2real_deployment}

At deployment, we keep the same observation normalization, group-wise action scaling, and stage-conditioned chunk horizons used during training. Sensing streams are timestamp-aligned, and the policy receives the most recent synchronized visual, tactile, and proprioceptive observations.

Stage transitions are triggered by geometric proximity and tactile onset. Let $d_{\mathrm{geo}}(t)$ be the distance from the knife to the pre-cut target region and let $\bar p_t$ be the aggregate tactile pressure. We use the gating rule
\begin{equation}
z_t
=
\begin{cases}
z^{\mathrm{app}},
&
d_{\mathrm{geo}}(t)>\epsilon_{\mathrm{app}},
\\[2mm]
z^{\mathrm{knife}},
&
d_{\mathrm{geo}}(t)\leq \epsilon_{\mathrm{app}}
\;\;\mathrm{and}\;\;
\bar p_t<p_{\mathrm{on}},
\\[2mm]
z^{\mathrm{cut}},
&
\bar p_t\geq p_{\mathrm{on}}.
\end{cases}
\label{eq:supp_stage_gate}
\end{equation}
If contact is lost for $K_{\mathrm{lost}}$ consecutive control steps,
\[
\bar p_t<p_{\mathrm{off}},
\qquad
p_{\mathrm{off}}<p_{\mathrm{on}},
\]
the controller exits tactile refinement and returns to knife positioning before attempting contact again. All executed commands pass through the same projection $\Pi_{\mathcal Q}$ in Eq.~\eqref{eq:supp_joint_update}, ensuring joint limits, rate limits, workspace bounds, and self-collision constraints are respected.
\section{Implementation Details}
\label{sec:impl_details}

\subsection{Two-Embodiment Setup Verification}
\label{subsec:two_embodiment_setup}

We evaluate DexNinja on two bimanual robot embodiments to verify that the real2sim2real pipeline is not tied to a single hardware configuration. The primary embodiment is used for the main real-world experiments, while the second embodiment is used to test embodiment-level robustness under a different arm--hand morphology. Across both settings, we keep the algorithmic pipeline unchanged: demonstrations are represented through object-centric knife trajectories, simulation data are generated in the same canonical object frame, and the learned policy keeps the same kinematic--tactile action decomposition. Only embodiment-specific calibration, robot kinematics, inverse kinematics, joint limits, and action scaling are changed.

\begin{table}[t]
\centering
\small
\begin{tabular}{p{0.22\linewidth}p{0.34\linewidth}p{0.34\linewidth}}
\hline
\textbf{Component} & \textbf{Embodiment A: primary setup} & \textbf{Embodiment B: transfer setup} \\
\hline
Robot arms & DR3/DR5 bimanual arm configuration & Dual Franka arm configuration \\
Dexterous hands & LinkerHand L10 dexterous hands & LinkerBot L20 dexterous hands \\
Tactile sensing & Pressure-based tactile sensors on both hands & Pressure-based tactile sensors on both hands \\
Vision & Fixed RGB-D camera (Orbbec 435Le) & Fixed RGB-D camera with recalibrated extrinsics \\
Role in evaluation & Main quantitative real-world experiments & Cross-embodiment qualitative verification \\
\hline
\end{tabular}
\caption{Two-embodiment hardware verification. Embodiment A is the primary DexNinja platform used for the main experiments. Embodiment B replaces the arms and hands while preserving the object-centric real2sim2real pipeline.}
\label{tab:two_embodiment_hardware}
\end{table}

Expert demonstrations are collected using data-collection gloves. Before training or retargeting, all sensor streams are timestamp-aligned and calibrated into the corresponding robot/world frame. We use the aligned demonstrations as supervision after standard filtering, calibration, and safety validation.

\begin{figure}[t]
    \centering
    \includegraphics[width=0.95\linewidth]{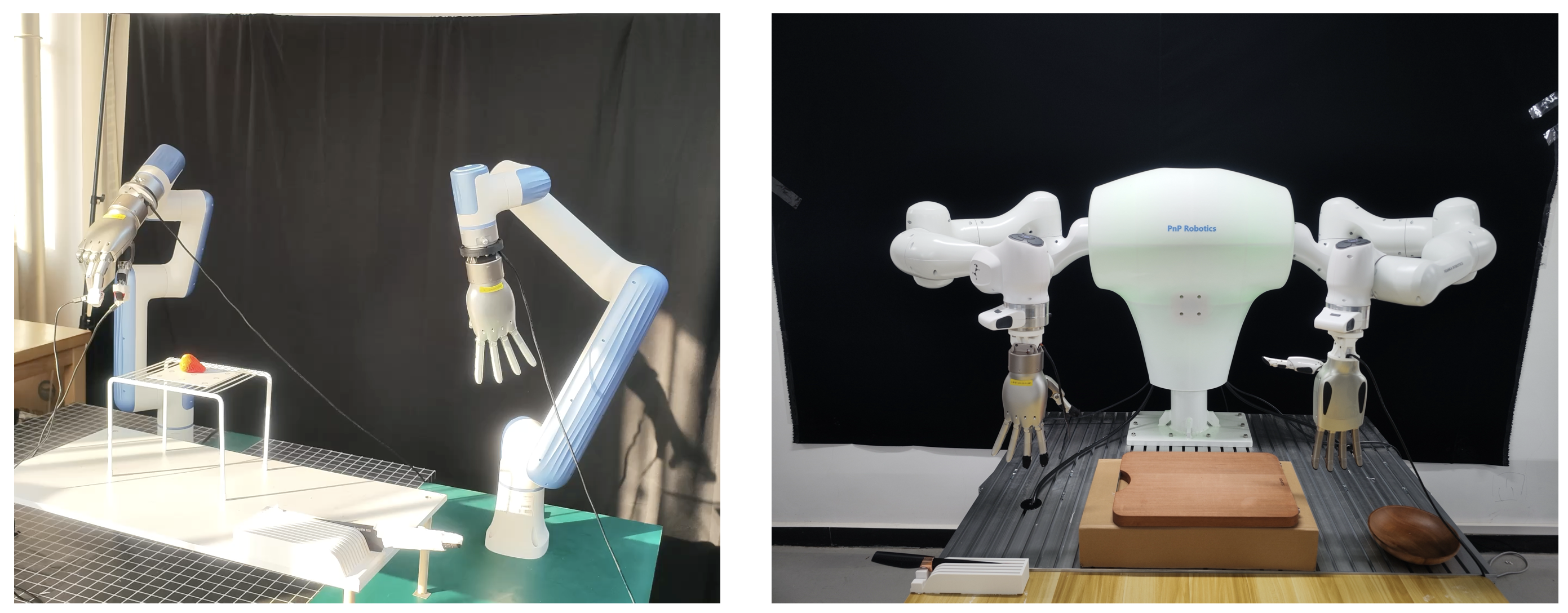}
    \caption{
    Two-embodiment setup verification.
    (a) Primary DexNinja setup with DR3/DR5 arms and LinkerHand L10 dexterous hands.
    (b) Embodiment-shift setup with Franka arms and LinkerBot L20 dexterous hands.
    The real2sim2real pipeline is kept unchanged at the algorithmic level; only robot-specific calibration, kinematic models, IK solvers, joint limits, and safety projections are updated.
    }
    \label{fig:two_embodiment_setup}
\end{figure}

This verification tests whether the proposed representation is portable across embodiments. Since the cutting motion is represented through the relative blade--object geometry in the canonical object frame, the same demonstrated skill can be transferred to a different robot by re-estimating the end-effector--knife transform and resolving the corresponding IK problem for the target embodiment.

\subsection{Embodiment-Specific Control and Safety Filters}
\label{subsec:embodiment_control_safety}

For both embodiments, we use the same high-level action decomposition as the main policy:
\[
a_t=
\left(
a^{\mathrm{kin}}_t,
a^{\mathrm{tac}}_t
\right).
\]
The kinematic component $a^{\mathrm{kin}}_t$ controls arm, wrist, and knife motion through embodiment-specific IK and joint-space tracking. The tactile component $a^{\mathrm{tac}}_t$ controls dexterous-hand contact regulation using pressure-based tactile feedback and proprioception.

Because the two embodiments have different kinematic chains, joint limits, and hand actuation ranges, the normalized policy outputs are mapped to executable joint commands through embodiment-specific scaling:
\begin{equation}
\Delta \hat{\mathbf q}^{(b)}_t
=
\mathbf{s}^{(b)}
\odot
\operatorname{clip}
\left(
\hat a_t,
-1,
1
\right),
\qquad
b\in\{\mathrm{A},\mathrm{B}\}.
\label{eq:supp_impl_embodiment_action_scaling}
\end{equation}
The commanded joint state is projected into the safe command set of the active robot:
\begin{equation}
\mathbf q^{(b)}_{t+1}
=
\Pi_{\mathcal Q^{(b)}}
\left(
\mathbf q^{(b)}_t
+
\Delta \hat{\mathbf q}^{(b)}_t
\right),
\label{eq:supp_impl_embodiment_projection}
\end{equation}
where $\Pi_{\mathcal Q^{(b)}}$ enforces the embodiment-specific joint limits, rate limits, workspace bounds, self-collision constraints, and environment-collision constraints.

During cutting, tactile feedback is downsampled and filtered before being passed to the dexterous-hand controller. This avoids impulsive high-frequency finger commands near contact. In practice, the hand controller runs at a lower rate than arm pose tracking and is used primarily to maintain stable grip, detect incipient slip, and prevent over-compression. Knife-pose targets that violate collision constraints, exceed joint limits, or approach kinematic singularities are discarded or repaired before execution.

The second embodiment therefore serves as a direct verification of the modularity of DexNinja: the object-centric motion template, visual--tactile observation format, and real2sim2real data-generation process remain unchanged, while only the robot-specific execution layer is swapped.

\section{Knowledge-Guided Data Augmentation}
\label{sec:supp_kgig}

We implement the knowledge-guided augmentation pipeline as a staged, loop-closure curation process that converts real-world captures into: (i) clean single-object reference images with transparent backgrounds, and (ii) validated 3D assets (3DGS or meshes) that pass multi-view consistency checks.

\paragraph{Inputs}
For each raw capture, we start from an RGB (or RGB-D) image $I$ that may contain clutter, occlusion, and background context.
A detector predicts the food category $y\in\mathcal{Y}$.
A summary agent extracts attribute-level descriptors $\mathbf{s}=\mathrm{Summarize}(I,y),$
including coarse shape (e.g., elongated vs.\ round), surface (e.g., smooth vs.\ bumpy),
color (e.g., pale green to deep red), and ripeness cues (e.g., glossiness, bruises).
We store $\mathbf{s}$ as a structured record to enforce cross-domain consistency (simulation and real)
and to condition subsequent generation.

\paragraph{Template-constrained description synthesis}
Raw captures often have partial views or quality issues (motion blur, occlusions).
Instead of directly using $I$ as a canonical reference, a generation agent produces a detailed text description
\[
p=\mathrm{Describe}(I,y,\mathbf{s}),
\]
formatted in a constrained template to improve controllability, e.g.,
\emph{``This is a \texttt{<adj>} \texttt{<object>} with \texttt{<features>}''},
where \texttt{<features>} enumerates extracted attributes in a fixed order (shape/surface/color/ripeness).

\paragraph{Image cleaning via inpainting and transparency}
Given $(I,p,\mathbf{s})$, we inpaint a single-object canonical view:
\[
\tilde{I}=\cradded{\mathrm{Inpaint}}(I,p,\mathbf{s}),
\]
enforcing: (i) one centered object, (ii) no hands/tools/background clutter, and (iii) transparent background (alpha matte).
The goal is to remove occluders and background while preserving instance attributes encoded by $\mathbf{s}$ and $p$.

\paragraph{Automatic quality control (QC) for 2D}
We filter generated images using a vision-language judge.
Given $(\tilde{I},y,\mathbf{s},p)$, the judge returns a decision $q_{\text{2D}}\in\{0,1\}$ and optional rejection reasons
(e.g., \emph{multiple objects}, \emph{background not transparent}, \emph{attribute mismatch}, \emph{artifacts}).
Rejected samples may be re-inpainted with stricter constraints derived from rejection reasons.

\paragraph{3D asset generation and multi-view validation}
For each retained clean image $\tilde{I}$, we generate a 3D representation using an image-to-3D backbone (e.g., Trellis~\cite{xiang2024structured}),
producing either a 3D Gaussian Splatting asset $\mathcal{G}$ or a mesh $\mathcal{M}$: $A \in \{\mathcal{G},\mathcal{M}\}=\mathrm{LiftTo3D}(\tilde{I}).$ We render $K$ views $\{\hat{I}_k\}_{k=1}^K$ from predefined camera poses and perform 3D quality check: $q_{\text{3D}}=\mathrm{JudgeMultiView}(\{\hat{I}_k\}_{k=1}^K,y,\mathbf{s}),$ checking (i) identity preservation across views, (ii) geometric plausibility (no missing lobes / melted structure),
and (iii) appearance stability (no view-dependent texture tearing).
Assets passing $q_{\text{3D}}=1$ are added to the category pool for simulation rollouts and local continuation around real anchors.
\cradded{Algorithm~\ref{alg:kgig} summarizes visual curation and simulation admission additionally requires metric-scale, geometry-validity, simulator-compatibility, and stable-initialization checks. Rollouts are retained only after IK, collision, and cutting-success checks, as described in the main paper.}

\begin{wrapfigure}{r}{0.45\linewidth}
    \centering
    \includegraphics[width=.9\linewidth]{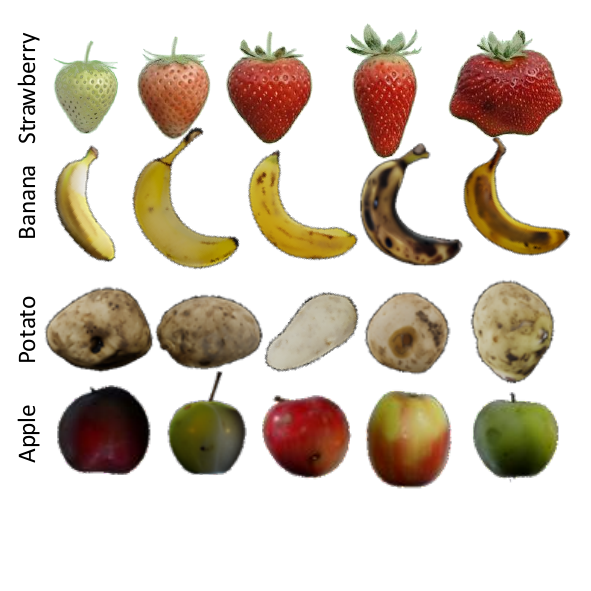}
    \caption{\textbf{Generated instance samples} from the knowledge-guided augmentation pipeline.}
    \label{fig:samples}
\end{wrapfigure}
\paragraph{Outputs}
The pipeline returns curated paired assets:
\[
\{(\tilde{I}_j,A_j,y_j,\mathbf{s}_j,p_j)\}_{j=1}^{N_{\text{keep}}},
\]
where each element is a transparent single-object image, a validated 3DGS/mesh, and its structured attributes.
These assets approximate a controllable neighborhood around real anchors for densification (Sec.~\cradded{\ref{sec:data augmentation}} of the main paper).

\begin{algorithm}[t]
\caption{Knowledge-guided instance generation with 2D/3D quality control}
\label{alg:kgig}
\small
\DontPrintSemicolon
\KwIn{Raw real-world captures $\{I_n\}_{n=1}^N$}
\KwOut{Curated assets $\mathcal{A}=\{(\tilde{I},A,y,\mathbf{s},p)\}$}
$\mathcal{A}\leftarrow\emptyset$\;
\For(\tcp*[f]{Process each capture}){$n=1$ \KwTo $N$}{
  $y \leftarrow \mathrm{Detect}(I_n)$\;
  $\mathbf{s} \leftarrow \mathrm{Summarize}(I_n,y)$\;
  $p \leftarrow \mathrm{Describe}(I_n,y,\mathbf{s})$\;
  $\tilde{I} \leftarrow \mathrm{Inpaint}(I_n,p,\mathbf{s})$\;

  $(q_{\text{2D}},r_{\text{2D}})\leftarrow \mathrm{Judge2D}(\tilde{I},y,\mathbf{s},p)$\;
  \If{$q_{\text{2D}}=0$}{
    $\tilde{I}\leftarrow \mathrm{ReInpaint}(\tilde{I},I_n,p,r_{\text{2D}})$\;
    $(q_{\text{2D}},r_{\text{2D}})\leftarrow \mathrm{Judge2D}(\tilde{I},y,\mathbf{s},p)$\;
  }
  \If{$q_{\text{2D}}=0$}{\textbf{continue}}

  $A \leftarrow \mathrm{LiftTo3D}(\tilde{I})$ \tcp*{$A$ is 3DGS or mesh}
  $\{\hat{I}_k\}_{k=1}^K \leftarrow \mathrm{RenderMultiView}(A)$\;

  $(q_{\text{3D}},r_{\text{3D}})\leftarrow \mathrm{JudgeMultiView}(\{\hat{I}_k\},y,\mathbf{s})$\;
  \If{$q_{\text{3D}}=0$}{\textbf{continue}}

  $\mathcal{A}\leftarrow \mathcal{A}\cup\{(\tilde{I},A,y,\mathbf{s},p)\}$\;
}
\Return{$\mathcal{A}$}\;
\end{algorithm}
\section{Additional Experiments}
\label{sec:addl_experiments}

This section provides additional real-world evaluations and ablations that complement the main experimental results. We focus on \cradded{four} questions:
(i) how DexNinja compares with the prior cutting baseline \textsc{RoboNinja}~\cite{xu2023roboninja};
(ii) how real-world performance changes as the number of simulated densification episodes increases;
\cradded{(iii)} how action-head factorization and adaptive action chunking affect performance; and
\cradded{(iv)} why tactile feedback is necessary for regulating contact rather than merely increasing grasp force.

Unless explicitly stated otherwise, we keep 
\cradded{the policy architecture}, the optimization hyperparameters, and the safety filters fixed. Thus, differences across experiments are primarily attributable to the ablated training data, test-time perturbation, or system component.

\subsection{Comparison to baseline}
\label{subsec:baseline_comp}

We first compare DexNinja with \textsc{RoboNinja} under the same real-world evaluation protocol. The comparison uses \cradded{40 trials per method}, with 10 trials per object category. As shown in \cref{tab:baseline_roboninja_vs_ours_avg}, DexNinja achieves $23/40$ successes ($57.5\%$), while \textsc{RoboNinja} achieves $12/40$ successes ($30.0\%$). This corresponds to a $+27.5$ percentage-point improvement and a $1.92\times$ relative gain in success rate.

\begin{table}[t]
  \centering
  \caption{Baseline comparison between RoboNinja  and our method on four objects.
  We report successes over 10 trials per object; \textbf{Avg.} aggregates over all 40 trials.}
  \label{tab:baseline_roboninja_vs_ours_avg}
  \begin{tabular}{lccccc}
    \toprule
    Method & Strawberry & Potato & Banana & Apple & Avg. \\
    \midrule
    RoboNinja (baseline) & 1/10 & 2/10 & 2/10 & 7/10 & 12/40 \\
    Ours                 & 6/10 & 5/10 & 9/10 & 3/10 & 23/40 \\
    \bottomrule
  \end{tabular}
\end{table}

The gains are distributed across multiple deformable foods: DexNinja improves over \textsc{RoboNinja} on Strawberry, Potato, and Banana. Apple remains a difficult category in this comparison, where the baseline obtains higher success. We attribute this residual gap to category-specific geometry, surface condition, and contact-regime variation that is not yet fully covered by the current densification distribution. This suggests a targeted augmentation direction: increasing the diversity of apple geometry, skin friction, cutting resistance, and initial contact configurations.

\paragraph{Baseline implementation.}
We implement \textsc{RoboNinja} as faithfully as possible from the publicly available description while matching our task setup and evaluation protocol. The original \textsc{RoboNinja} setting assumes a rigidly attached knife, which removes grasp slippage and simplifies the tool--hand interaction. Since DexNinja requires the dexterous hand to grasp and manipulate the knife, we rigidly attach the knife when evaluating \textsc{RoboNinja} to preserve the baseline assumption.

The exact force-sensor model and calibration used in the original \textsc{RoboNinja} implementation were not fully recoverable from the public materials. We therefore approximate the required contact feedback with a scalar contact/force estimate derived from our tactile readings. This provides a comparable contact-aware signal, but it may not exactly reproduce the original sensor noise, bandwidth, or calibration. The baseline results should therefore be interpreted as a protocol-matched reproduction with an approximate contact-sensing interface.

\subsection{Simulation scaling and generation cost}
\label{subsec:sim_scaling_cost}

We next evaluate how the scale of simulation-based densification affects real-world performance. We vary the number of simulated episodes added to a fixed set of real teleoperated demonstrations, while keeping the architecture, optimization hyperparameters, action representation, and evaluation protocol unchanged.

\begin{table}[t]
  \centering
  \caption{Scaling law of simulation augmentation for real-world cutting. We train with a fixed real dataset and add $N$ simulated episodes; values are real-world \textbf{success} (25 trials per food), and Avg. aggregates 100 trials.}
  \label{tab:scaling_sim_episodes_real}
  \begin{tabular}{lccccc}
    \toprule
    \textbf{+Sim episodes $N$} & \textbf{Strawberry} & \textbf{Banana} & \textbf{Potato} & \textbf{Apple} & \textbf{Avg.} \\
    \midrule
    10  & 9/25  & 14/25 & 8/25  & 5/25  & 36/100 \\
    50  & 12/25 & 18/25 & 12/25 & 8/25  & 50/100 \\
    100 & \pp{14/25} & \pp{20/25} & \pp{13/25} & \pp{9/25}  & \pp{56/100} \\
    300 & \best{16/25} & \best{22/25} & \best{15/25} & \best{10/25} & \best{63/100} \\
    \bottomrule
  \end{tabular}
\end{table}

As shown in \cref{tab:scaling_sim_episodes_real}, increasing the number of simulated episodes generally improves real-world success. This trend supports the role of simulation as a data densification engine: additional simulated rollouts expand coverage over geometry, contact, and material variations that are difficult to collect exhaustively from real demonstrations.

The improvement is not strictly linear. We observe diminishing returns once common interaction modes are sufficiently covered, and the finite number of real-world trials introduces evaluation variance. In our current implementation, generating one single-cut simulated episode—including deformable cutting and tactile synthesis—requires approximately \cradded{$20\,\mathrm{GB}$} of GPU memory and 10 minutes on an NVIDIA H200 GPU. Thus, generating $N$ single-cut episodes has a nominal cost of about $10N$ minutes, or $N/6$ GPU-hours, before parallelization. Multi-cut simulations scale similarly. A two-cut episode requires roughly \cradded{$30\,\mathrm{GB}$} of GPU memory and 20 minutes; more generally, an $M$-cut episode requires approximately \cradded{$(10+10M)\,\mathrm{GB}$} of GPU memory and $M/6$ GPU-hours. Therefore, for $N$ episodes with $M$ cuts each, the nominal cost is approximately $NM/6$ GPU-hours before parallelization. Given this near-linear generation cost, the scaling results suggest that future data generation should prioritize diversity and coverage of interaction modes, rather than simply increasing the number of rollouts.





\subsection{System-level ablations}
\label{subsec:system_ablation}

We further ablate two design choices used by the DexNinja policy: action-head factorization and adaptive action chunking. The action-head ablation compares a universal action head, which predicts the full arm--hand command with a shared output head, against separate heads for the kinematic and tactile/contact components. The chunking ablation compares a fixed chunk size of 64 against the stage-conditioned adaptive chunking strategy used by the full system.

\begin{table}[t]
\centering
\caption{System-level ablation study on head design and action chunking across objects. Metric: task success rate (successes / 10 trials).}
\label{tab:ablation_merged}
\resizebox{\linewidth}{!}{
\begin{tabular}{lcccc}
\hline
Object & Universal Head & Separate Heads & Fixed Action Chunk & Full System \\
\hline
Apple      & 2/10 & 4/10 & 1/10 & 5/10 \\
Banana     & 4/10 & 7/10 & 6/10 & 7/10 \\
Strawberry & 2/10 & 5/10 & 2/10 & 7/10 \\
Potato     & 3/10 & 5/10 & 4/10 & 5/10 \\
\hline
Average    & 11/40 & 21/40 & 13/40 & 24/40 \\
\hline
\end{tabular}
}
\end{table}

As summarized in \cref{tab:ablation_merged}, the universal action head underperforms the separate-head design. This supports the policy factorization in the main paper: the kinematic component $a^{\mathrm{kin}}_t$ handles large-scale arm, wrist, and tool motion, while the tactile component $a^{\mathrm{tac}}_t$ handles local contact-sensitive hand control. Combining both action groups in a single head creates a scale mismatch between arm motion and finger motion, which makes optimization less stable.

The fixed-chunk ablation also performs worse than the full system. A single chunk horizon is poorly matched to the temporal structure of dexterous cutting: approach and knife positioning benefit from longer, smoother chunks, whereas contact-rich cutting requires shorter chunks and more frequent replanning. The full system combines separate action heads with adaptive stage-conditioned chunking and achieves the best aggregate performance.

\subsection{Tactile regulation versus forceful grasping}
\label{subsec:tactile_regulation}

Tactile sensing is not used simply to grasp more tightly. Instead, it helps maintain the system in the narrow contact regime between slip and over-compression. This distinction is important for deformable foods: increasing the normal force may suppress slip, but it can also damage the object, shift the cutting target, or cause excessive deformation before the knife completes the cut.

We include a guava contact comparison in \cref{fig:tactile_important}. The nominal-contact case shows localized tactile activation on the thumb and index finger, while the high-force case produces a broader and stronger tactile response, corresponding to visible compression. This qualitative result supports the design of the tactile policy $\pi_\theta$: the contact controller should regulate pressure and shear, rather than monotonically increasing grasp force.

\begin{figure}[htbp]
    \centering
    \includegraphics[width=0.95\linewidth]{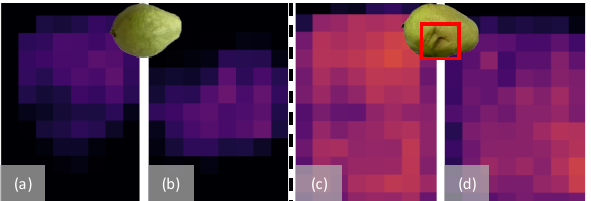}
    \caption{
    Tactile responses under nominal and high-force contact.
    (a,b) Thumb and index tactile maps under nominal contact with an undeformed guava.
    (c,d) Thumb and index tactile maps under high-force contact with a compressed guava.
    The comparison illustrates why tactile feedback is needed to regulate contact within the slip--over-compression regime.
    }
    \label{fig:tactile_important}
\end{figure}

\subsection{Summary of additional findings}
\label{subsec:addl_summary}

The additional experiments support \cradded{three} conclusions. First, under the same real-world protocol, DexNinja outperforms the \textsc{RoboNinja} baseline in aggregate success, supporting the benefit of real2sim2real densification for contact-rich dexterous cutting. Second, increasing simulation data generally improves real-world success, but the gains saturate, motivating diversity-aware simulation rather than only larger episode counts. \cradded{Third}, the system-level ablations validate the main policy design: separate kinematic and tactile action heads, together with adaptive action chunking, are important for stable real-world execution.
\section{Qualitative Results}
\label{sec:qual_results}

We include representative real-world executions in Fig.~\ref{fig:supp_qualitative}, with 2 different embodiment settings \cradded{to illustrate cross-embodiment deployment}.
Background colors denote object type: orange (banana), \cradded{pink (strawberry), red (apple)}, and yellow (potato).

\begin{figure*}[t]
    \centering
    \includegraphics[width=\linewidth]{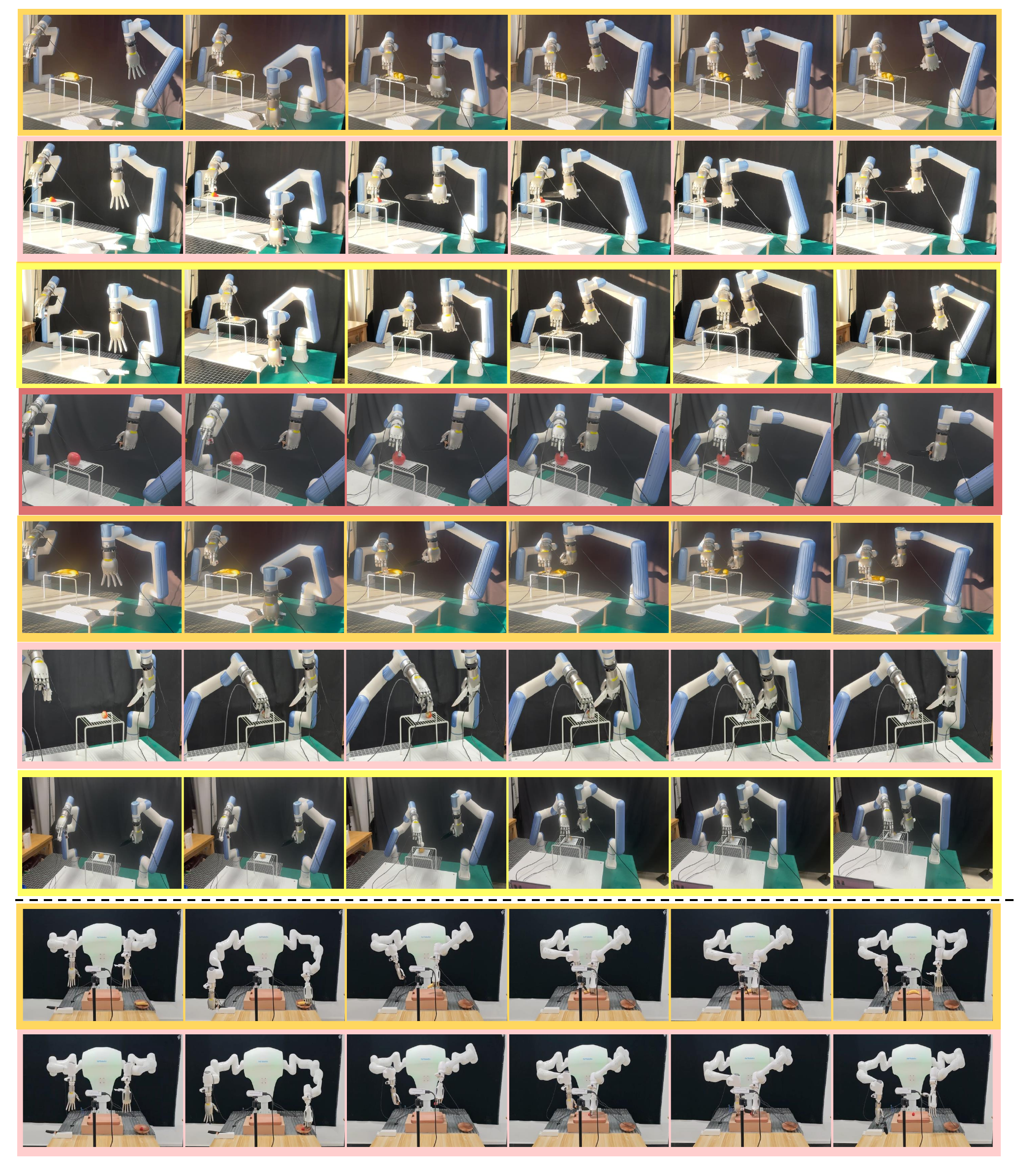}
    \caption{
\textbf{Qualitative real-world deployment results with 2 different embodiment settings.}
Representative executions across food categories.
Background colors denote the object type:
\textcolor{orange}{orange} (banana),
\textcolor{pink}{pink} (strawberry),
\textcolor{red}{red} (apple), and
\textcolor{yellow}{yellow} (potato).
}
    \label{fig:supp_qualitative}
\end{figure*}

\begin{figure*}[t]
    \centering
    \includegraphics[width=\linewidth]{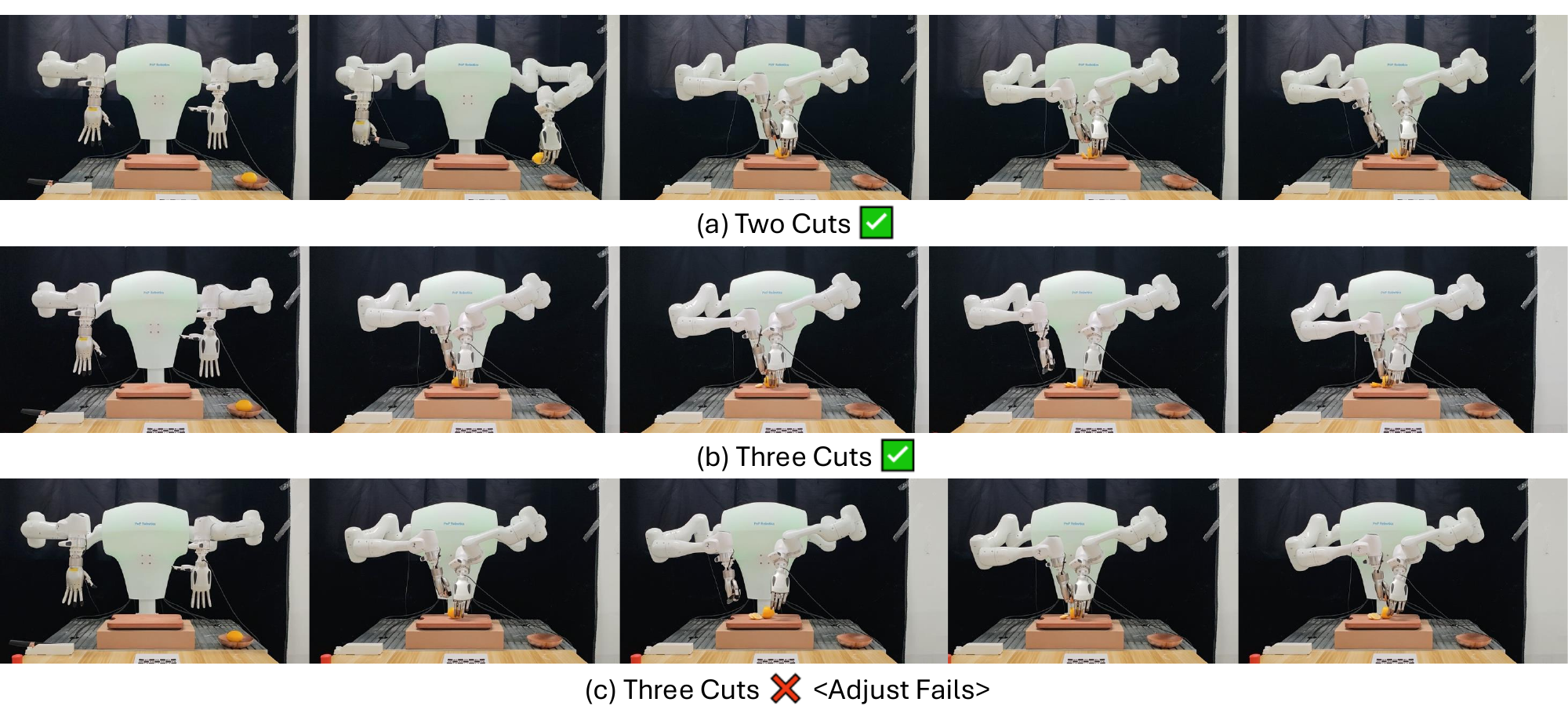}
    \caption{
\textbf{Qualitative Multi-cuts Experiments results.}
}
    \label{fig:supp_qualitative_multi_cut}
\end{figure*}

\paragraph{Knife initialization and handover.}
Knife pre-grasping is initialized using DexGraspNet~2.0~\cite{zhang2024dexgraspnet20learninggenerative}.
Concretely, we use the generative grasp prior to propose stable handle grasps and an associated end-effector
pre-grasp pose, then execute a short approach to bring the knife into a consistent, repeatable starting
configuration before closed-loop control begins.
This initialization serves two purposes: (i) it reduces the burden on the downstream policy by standardizing
the initial knife pose (position, yaw/pitch/roll, and approach height), and (ii) it improves safety and
repeatability by ensuring the knife starts outside the contact region with sufficient clearance from the
board and the object.
After the pre-grasp is reached, our policy takes over to refine alignment, initiate contact, and complete
the slice.

\paragraph{Multi-cut Experiment Results.}
We further evaluate real-world multi-cut execution, as shown in Fig.~\ref{fig:supp_qualitative_multi_cut}. In the two-cut setting, the execution procedure remains similar to the single-cut case, and the additional cut does not substantially change the interaction sequence. In contrast, the three-cut setting introduces a longer-horizon manipulation problem that requires repeated fruit adjustment and re-alignment before subsequent cuts. We observe that failures occur primarily during this fruit-adjustment stage, where accumulated pose errors and contact disturbances can prevent reliable re-positioning for the next cut.

\paragraph{Phase structure in real rollouts.}

During approach, the knife translates to the estimated pre-cut region while keeping a conservative
clearance and a stable orientation.
In contact initiation, the policy performs small, corrective motions (typically millimeter-scale
translations and a slight pitch adjustment) to settle the blade against the surface without inducing large
lateral forces that would slip or topple the object.
Finally, during slicing, the knife follows a smooth, monotonic progression through the object; we
often observe minor orientation compensation to maintain an effective cutting angle as the contact state
changes (e.g., as the blade transitions from skin contact to deeper penetration).

Qualitatively, different food categories induce distinct contact regimes that are reflected in the knife
trajectory and pose evolution.
For \textbf{banana} (soft, high compliance), successful trials tend to use a shallower entry and shorter,
more continuous slicing strokes to avoid excessive deformation that would cause the target region to drift.
For \textbf{strawberry} (delicate surface and easily damaged interior), the rollouts emphasize gentle
contact initiation and reduced tangential loading; the knife trajectory typically prioritizes stable
alignment before committing to depth, which helps preserve \emph{integrity} (i.e., limiting damage to the
intended cut).
For \textbf{potato} (stiff, higher cutting resistance), we observe slower progression with more pronounced
maintenance of blade orientation, consistent with the need to sustain cutting pressure while avoiding sudden
jumps that could lead to chatter or board collision.

\paragraph{Common qualitative failure modes.}
When failures occur, they are visually attributable to a small set of recurring issues:
(i) \emph{initial misalignment} (the blade contacts off-center, leading to a biased cut or early slip),
(ii) \emph{premature contact} during approach (insufficient clearance causes the knife to bump the object
and disturb its pose), and
(iii) \emph{late-stage jamming or excessive friction} (the knife stalls or deviates as the contact patch
grows). (iv) \emph{multi-cut re-alignment failure} in longer-horizon trials. In particular, while two-cut executions
remain close to the single-cut setting, three-cut executions require repeated fruit adjustment and
re-alignment between cuts. Failures in these trials often occur during the adjustment stage, where
accumulated pose errors or contact disturbances prevent the fruit from being reliably positioned for
the next cut.

\clearpage
\endgroup

\end{document}